\documentclass[10pt,twocolumn,letterpaper]{article}

\usepackage{iccv}              

\usepackage{makecell}
\usepackage{multicol}
\usepackage{multirow}
\usepackage[table]{xcolor}

\newcommand{\first}[1]{\ifblank{#1}{\textbf{bold}}{\textbf{#1}}}

\definecolor{roof}{rgb}{0.47, 0.04, 0.13}
\definecolor{dirt}{rgb}{0.71, 0.65, 0.71}
\definecolor{pavedmotor}{rgb}{0.5, 0.25, 0.5}
\definecolor{river}{rgb}{0.68, 0.85, 0.9}
\definecolor{pool}{rgb}{0.0, 0.31, 0.39}
\definecolor{bridge}{rgb}{0.59, 0.39, 0.39}
\definecolor{container}{rgb}{0.98, 0.67, 0.12}
\definecolor{airstrip}{rgb}{0.32, 0.0, 0.32}
\definecolor{traffic}{rgb}{0.4, 0.4, 0.61}
\definecolor{greenfield}{rgb}{0.42, 0.56, 0.14}
\definecolor{wildfield}{rgb}{0.82, 0.71, 0.55}
\definecolor{solar}{rgb}{0.86, 0.86, 0.0}
\definecolor{umbrella}{rgb}{0.6, 0.6, 0.6}
\definecolor{transparent}{rgb}{0.0, 0.0, 0.35}
\definecolor{carpark}{rgb}{0.98, 0.67, 0.63}
\definecolor{pavedwalk}{rgb}{0.96, 0.14, 0.91}
\definecolor{sedan}{rgb}{0.0, 0.0, 0.56}
\definecolor{truck}{rgb}{0.0, 0.0, 0.27}

\definecolor{iccvblue}{rgb}{0.21,0.49,0.74}
\usepackage[pagebackref,breaklinks,colorlinks,allcolors=iccvblue]{hyperref}

\def\paperID{9316} 
\def\confName{ICCV}
\def\confYear{2025}

\title{UAVScenes: A Multi-Modal Dataset for UAVs}

\author{
\makebox[\textwidth][c]{%
Sijie Wang\textsuperscript{*1} \quad
Siqi Li\textsuperscript{*1} \quad
Yawei Zhang\textsuperscript{*1} \quad
Shangshu Yu\textsuperscript{*2} \quad
Shenghai Yuan\textsuperscript{*1} \quad
Rui She\textsuperscript{*3}
}\\
\makebox[\textwidth][c]{%
Quanjiang Guo\textsuperscript{4} \enspace
JinXuan Zheng\textsuperscript{1} \enspace
Ong Kang Howe\textsuperscript{1} \enspace
Leonrich Chandra\textsuperscript{1} \enspace
Shrivarshann Srijeyan\textsuperscript{1} 
}\\
\makebox[\textwidth][c]{%
Aditya Sivadas\textsuperscript{1} \enspace
Toshan Aggarwal\textsuperscript{1} \enspace
Heyuan Liu\textsuperscript{1} \enspace
Hongming Zhang\textsuperscript{1} \enspace
Chujie Chen\textsuperscript{1} \enspace
Junyu Jiang\textsuperscript{1}
}\\
\makebox[\textwidth][c]{%
Lihua Xie\textsuperscript{1} \quad
Wee Peng Tay\textsuperscript{1}
}\\
\textsuperscript{1}Nanyang Technological University \\ 
\textsuperscript{2}School of Computer Science and Engineering, Northeastern University, Shenyang 110819, China \\ 
\textsuperscript{3}Beihang University \\
\textsuperscript{4}University of Electronic Science and Technology of China \\
{\tt\small wang1679@e.ntu.edu.sg, shyuan@ntu.edu.sg} \\
}

\begin{document}
\maketitle

\begin{abstract}
Multi-modal perception is essential for unmanned aerial vehicle (UAV) operations, as it enables a comprehensive understanding of the UAVs' surrounding environment. However, most existing multi-modal UAV datasets are primarily biased toward localization and 3D reconstruction tasks, or only support map-level semantic segmentation due to the lack of frame-wise annotations for both camera images and LiDAR point clouds. This limitation prevents them from being used for high-level scene understanding tasks. To address this gap and advance multi-modal UAV perception, we introduce UAVScenes, a large-scale dataset designed to benchmark various tasks across both 2D and 3D modalities.
Our benchmark dataset is built upon the well-calibrated multi-modal UAV dataset MARS-LVIG, originally developed only for simultaneous localization and mapping (SLAM). We enhance this dataset by providing manually labeled semantic annotations for both frame-wise images and LiDAR point clouds, along with accurate 6-degree-of-freedom (6-DoF) poses. These additions enable a wide range of UAV perception tasks, including segmentation, depth estimation, 6-DoF localization, place recognition, and novel view synthesis (NVS). 
Our dataset is available at \url{https://github.com/sijieaaa/UAVScenes}
\renewcommand{\thefootnote}{\fnsymbol{footnote}}
\footnotetext[1]{The first six authors contribute equally: Sijie Wang, Siqi Li, Yawei Zhang, Shangshu Yu, Shenghai Yuan, and Rui She.}
\renewcommand{\thefootnote}{\arabic{footnote}} 
\end{abstract}

\section{Introduction}

With the expansion of the low-altitude aerial economy \cite{kyrkou2019deep, shakhatreh2019unmanned, mohsan2022towards, jiang20236g_lowaltitude_economy}, UAVs have become indispensable for aerial taxi services \cite{cao2022direct,cao2023path,cao2023neptune}, low-altitude logistics \cite{betti2024uav_delivery}, agriculture \cite{mogili2018review_agriculture}, inspection \cite{xu2024cost,lyu2023vision,lyu2022structure,cao2020online}, and emergency response \cite{jiang20236g_lowaltitude_economy}. Unlike ground vehicles, UAVs can operate above ground constraints, addressing limitations in current urban systems.

\begin{figure}[!tb]
\vspace{-4pt}
\centering
\includegraphics[width=0.99\linewidth]{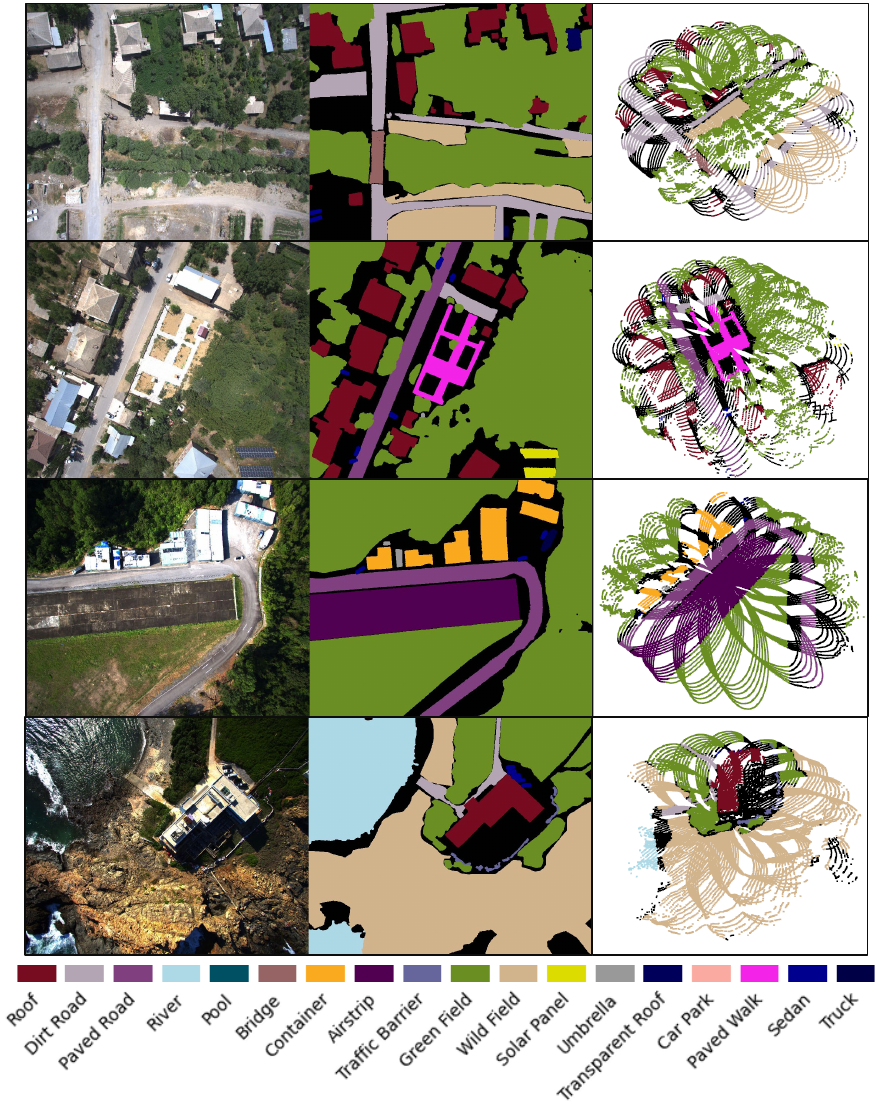}
\vspace{-8pt}
\caption{Visualization of frame-wise image and LiDAR point cloud annotations of the proposed UAVScenes dataset.}
\label{fig:image_2d_3d_anno}
\vspace{-8pt}
\end{figure}

Yet, as UAVs tackle increasingly sophisticated tasks in diverse urban settings, they require training on high-quality datasets for reliable perception. Despite the potential of a growing low-altitude economy, most existing datasets focus on single-modality camera data \cite{lyu2020uavid, du2018unmanned_uavdt, rahnemoonfar2021floodnet, hsieh2017drone_carpk, nigam2018ensemble_aeroscapes, burri2016euroc, xu2024uav_uavvisloc}. Although cameras provide rich texture information, they cannot capture the vital 3D spatial data required for more comprehensive and robust scene understanding.

Recent perception advances enable UAVs to carry lightweight 3D sensors, such as depth cameras, which work well for close-range tasks but remain limited in broader applications \cite{alexovivc20233d_depthcamera_uav,wang2020uav_depthcamera,xu2023real_uavrgbd}. In contrast, 3D LiDAR offers long-range depth information and, when combined with cameras, provides multi-modal perception that substantially enhances UAV capabilities. While recent datasets \cite{nguyen2022ntu_viral,zhu2023graco,dhrafani2024firestereo,thalagala2024munfrl,xiong2024gauuscene,xiong2024gauu_gauuscenev2,li2024mars_lvig} integrate cameras and LiDAR for richer 3D data, most focus on SLAM or 3D reconstruction and lack annotations for broader UAV tasks. Other multi-modal datasets \cite{kolle2021hessigheim_h3d,lin2022capturing_urbanscene3d} only label 3D maps, limiting support for frame-wise tasks like image and LiDAR point cloud semantic segmentation.

In summary, despite advancements in UAV datasets, existing multi-modal UAV datasets either focus on SLAM~\cite{nguyen2021viral,yin2024outram} and 3D reconstruction~\cite{deng2024incremental,deng2024compact} or only label 3D maps~\cite{kolle2021hessigheim_h3d,lin2022capturing_urbanscene3d}. None provides full semantic annotations for frame-wise geo-referenced images and LiDAR data. This gap limits their utility for real-time aerial scene understanding~\cite{yang2022overcoming}, navigation~\cite{er2013development}, and precision operations~\cite{li2023drone_delivery,betti2024uav_delivery}.


To address this gap in UAV perception research, we present a large-scale annotated multi-modal benchmark dataset, \textbf{UAVScenes}. Built upon the MARS-LVIG dataset~\cite{li2024mars_lvig}, originally designed for SLAM, UAVScenes includes semantic annotations for frame-wise camera images and LiDAR point clouds (see \cref{fig:image_2d_3d_anno}), along with accurate 6-degree-of-freedom (6-DoF) poses and reconstructed 3D maps. It supports a wide range of multi-modal UAV perception tasks, including semantic segmentation, depth estimation, localization, and novel view synthesis (NVS). Our main contributions are as follows:
\begin{itemize}
\item We present UAVScenes, a comprehensive multi-modal dataset for UAV perception that provides robust semantic scene understanding for both images and LiDAR point clouds, along with accurate 6-DoF poses and reconstructed 3D maps.
\item Our dataset features over 120k frames with semantic annotations for images and LiDAR point clouds, surpassing the scale of most existing UAV research datasets.
\item We conduct extensive benchmarking and evaluation of state-of-the-art (SOTA) methods on our dataset, establishing it as a wide-ranging UAV perception benchmark that supports at least six distinct tasks.
\end{itemize}

\section{Related Work}

In this section, we summarize existing autonomous driving and UAV datasets, discussing their sensor modalities, task coverage, and environmental constraints. We then highlight their limitations in multi-modality and semantic labeling. By comparison, UAVScenes is designed to address these gaps by offering robust multi-modal coverage and frame-wise semantic annotations.

\subsection{UAV Datasets}

UAV perception datasets have become increasingly important due to the unique challenges posed by aerial perspectives. Over the years, various UAV datasets \cite{fonder2019midair,du2018unmanned_uavdt,hsieh2017drone_carpk,yan2022crossloc,wu2024uavd4l,sun2022drone_dronevehicle,nguyen2022ntu_viral,lin2022capturing_urbanscene3d,xiong2024gauu_gauuscenev2,li2024mars_lvig} have been proposed, each contributing to different aspects of UAV perception. 

\begin{table*}[!htb]
\centering
\scriptsize
\begin{tabular}{l | c | ccccccc}
\toprule
Dataset &  Modality  &  LiDAR Type  & \makecell{6-DoF  \\ Pose}    & \makecell{\#Real Camera Frames \\ with Frame-wise Anno. (type)}     &  \makecell{\#Real LiDAR Frames \\ with Frame-wise Anno. }     & \makecell{Multiple \\ Traversals}  & \makecell{3D \\ Map}  \\
\midrule
Mid-Air \cite{fonder2019midair}                   & Simulation     & no real LiDAR & \checkmark & no real camera & no real LiDAR  & \checkmark & \checkmark (with anno.)   \\
TartanAir \cite{wang2020tartanair}                & Simulation     & no real LiDAR & \checkmark & no real camera & no real LiDAR  & \checkmark & \checkmark (with anno.)    \\
University-1652 \cite{zheng2020university1652}    & Google Earth   & no real LiDAR & - & no real camera & no real LiDAR & - & - \\
SynDrone \cite{rizzoli2023syndrone}               & Simulation     & no real LiDAR & - & no real camera & no real LiDAR & - & - \\
\midrule
UAVDT \cite{du2018unmanned_uavdt}                 & C &-&-& 80k (bbox) & - & - & - \\
VisDrone \cite{zhu2018vision_visdrone}            & C &-&-& 40k (bbox) & - & - & - \\
CARPK \cite{hsieh2017drone_carpk}                 & C &-&-& 1k (bbox)  & - & - & - \\
Semantic Drone \cite{semantic_drone_dataset}      & C &-&-& 0.6k (mask) & - & - & - \\
Aeroscapes \cite{nigam2018ensemble_aeroscapes}    & C &-&-& 3k (mask)  & - & - & - \\
UAVid \cite{lyu2020uavid}                         & C &-&-& 0.3k (mask) & - & - & - \\
FloodNet \cite{rahnemoonfar2021floodnet}  & C &-&-& 9k (mask)  & - & - & - \\
CrossLoc \cite{yan2022crossloc}                   & C &-&\checkmark& 7k (mask) & - & \checkmark & \checkmark (with anno.)\\
ALTO \cite{cisneros2022alto}                      & C &-&-& -  & -  & -  & - \\
STPLS3D \cite{chen2022stpls3d}                      & C &-&-& -  & -  & -  & \checkmark (with anno.) \\
VDD \cite{cai2023vdd}                             & C &-&-& 0.4k (mask)        & -  & -  & - \\
SUES-200 \cite{zhu2023sues}                       & C &-&-& -           & -  & \checkmark & -\\
UAV-VisLoc \cite{xu2024uav_uavvisloc}             & C &-&-& -           & -  & \checkmark & - \\
HazyDet \cite{feng2024hazydet}                    & C &-&-& 12k (bbox)         & - & - & -\\
UAVD4L \cite{wu2024uavd4l}                        & C &-&\checkmark     & -  & - & - & \checkmark\\
\midrule
Hessigheim 3D  \cite{kolle2021hessigheim_h3d}     & C+L & RIEGL VUX-1LR & - & - & - & - & \checkmark (with anno.)\\
Drone Vehicle \cite{sun2022drone_dronevehicle}    & C+IR & - & -  & 57k (bbox) & - & - & -\\
NTU VIRAL \cite{nguyen2022ntu_viral}              & C+L   & 2$\times$3D-Ouster-16 & \checkmark & - & -  & \checkmark & \checkmark\\
UrbanScene3D \cite{lin2022capturing_urbanscene3d} & C+L   & Trimble-X7           & \checkmark & - & -  & \checkmark & \checkmark (with anno.)\\
GrAco \cite{zhu2023graco}                         & C+L   & Velodyne-16          & - & - & -  & \checkmark & -         \\
GauU-Scene V2 \cite{xiong2024gauu_gauuscenev2}    & C+L   & DJI-L1*              & \checkmark & - & - & \checkmark & \checkmark\\
FIReStereo \cite{dhrafani2024firestereo}          & C+L   & Velodyne-16          & - & - & - & - & - \\
MUN-FRL \cite{thalagala2024munfrl}                & C+L   & Velodyne-16          & \checkmark & - & - & \checkmark & - \\
MARS-LVIG \cite{li2024mars_lvig}                  & C+L   & DJI-L1* + Livox-Avia & - & - & -  & \checkmark & \checkmark\\
\midrule
UAVScenes (ours)                                  & C+L   & Livox-Avia & \checkmark & \first{120k (mask)}  & \first{120k}  & \checkmark & \checkmark (with anno.) \\               
\bottomrule
\end{tabular}
\vspace{-4pt}
\caption{Comparison of UAV datasets. Our dataset is the only one offering frame-wise annotations for both LiDAR and camera data on real scenes. We only count frame-wise annotations of real data, excluding synthetic or rendered data. ``C'' represents visible cameras, ``L'' represents LiDARs, and ``IR'' represents infrared cameras. ``-'' indicates that the dataset does not support this feature. ``*'' means that the DJI-L1 sensor produces encrypted point clouds, so per-frame LiDAR cannot be accessed. ``bbox'' denotes bounding boxes, and ``mask'' denotes semantic or instance masks.}
\label{tab:uav_datasets}
\vspace{-8pt}
\end{table*}


\noindent \textbf{Synthetic Datasets:} Synthetic UAV datasets are primarily generated using simulation tools or platforms like Google Earth and CARLA \cite{Dosovitskiy17_carla}. Typical datasets like Mid-Air \cite{fonder2019midair}, TartanAir \cite{wang2020tartanair}, University-1652 \cite{zheng2020university1652}, and SynDrone \cite{rizzoli2023syndrone} are used for various tasks. Mid-Air and TartanAir offer large-scale synthetic images and LiDAR-type data in unstructured environments. University-1652 features synthetic aerial images with satellites and ground views, providing the view when flying around the target. SynDrone offers semantic annotations for both synthetic LiDAR and camera.
These datasets provide sufficient synthetic drone-view data for localization or scene understanding tasks.



\noindent \textbf{{Camera-Only Datasets:}} Beyond synthetic datasets, early real-world UAV datasets predominantly consist of visual camera-only modality due to the high cost of sensors and the relative immaturity of fusion technologies. These datasets provide visual imagery typically used for camera-based tasks. For example, some datasets \cite{du2018unmanned_uavdt,zhu2018vision_visdrone,hsieh2017drone_carpk,semantic_drone_dataset,nigam2018ensemble_aeroscapes,lyu2020uavid,rahnemoonfar2021floodnet,yan2022crossloc,cai2023vdd,feng2024hazydet} include semantic or object annotations, supporting tasks such as semantic segmentation and object detection. Additionally, other datasets \cite{cisneros2022alto,zhu2023sues,xu2024uav_uavvisloc,wu2024uavd4l} provide location data for each image, which can be used to benchmark localization and place recognition models.

However, these datasets lack the 3D LiDAR modality, limiting their application in 3D scene understanding and high-precision multi-modal fusion tasks.

\noindent \textbf{Multi-Modal Datasets:} With the advancement of sensor technology, an increasing number of multi-modal UAV datasets \cite{kolle2021hessigheim_h3d,sun2022drone_dronevehicle,nguyen2022ntu_viral,lin2022capturing_urbanscene3d,zhu2023graco,xiong2024gauuscene,xiong2024gauu_gauuscenev2,dhrafani2024firestereo,thalagala2024munfrl,li2024mars_lvig} have emerged in recent years.

The H3D dataset~\cite{kolle2021hessigheim_h3d} provides annotations on 3D maps reconstructed by LiDAR and camera data. However, it does not contain frame-wise annotations, which limits its applicability for frame-wise perception evaluation.  

The Drone Vehicle dataset~\cite{sun2022drone_dronevehicle} enhances drone surveillance with labeled imagery for object detection and tracking. It also features infrared capabilities for visibility in low-light conditions. However, the absence of LiDAR restricts its use in 3D scene understanding and high-precision localization.
NTU VIRAL~\cite{nguyen2022ntu_viral} is a dataset designed for UAV SLAM and includes camera and LiDAR data. It enables research in tasks like place recognition, 3D mapping, and localization. However, it is collected in indoor and small-scale outdoor environments, limiting its application for large-scale scene understanding.
The UrbanScene3D dataset~\cite{lin2022capturing_urbanscene3d} provides high-resolution imagery and LiDAR data from an urban setting, offering capabilities for 3D scene segmentation \cite{yuan2014autonomous} and localization for UAVs. However, it only offers semantic information on the reconstructed 3D map without considering frame-wise LiDAR point clouds, which prevents benchmarking frame-wise camera \cite{li2025airswarm} and LiDAR \cite{lou2025qlio} scene parsing \cite{cai2025bev}.
GrAco~\cite{zhu2023graco} and FIReStereo~\cite{dhrafani2024firestereo} do not offer 6-DoF poses and focus on 3-DoF localization and stereo estimation \cite{wang2017heterogeneous}, respectively.
The GauU-Scene datasets~\cite{xiong2024gauuscene,xiong2024gauu_gauuscenev2} collect UAV camera and LiDAR data in various urban environments and provide geo-aligned 3D maps. However, GauU-Scene uses DJI-L1 LiDAR, a closed-source sensor with encrypted point cloud data, hindering frame-wise LiDAR perception.
MUM-FRL~\cite{thalagala2024munfrl} is equipped only with a short-range LiDAR, resulting in substantial undetected point cloud data on the ground due to the high flying altitude.
The MARS-LVIG~\cite{li2024mars_lvig} dataset stands out by providing multi-modal data across diverse scenarios, including multiple traversals through towns, valleys, airports, and islands. Additionally, it features a synchronized camera-LiDAR suite, ensuring well-aligned images and point clouds.

However, these existing multi-modal UAV datasets lack frame-wise annotations for both images and LiDAR point clouds, limiting their utility for benchmarking advanced multi-modal perception tasks. 


The UAVScenes dataset aims to fill this gap by providing comprehensive semantic annotations for both frame-wise images and LiDAR data. Additionally, it includes accurate 6-DoF poses and reconstructed point cloud maps, enabling a wide range of tasks such as segmentation, depth estimation, 6-DoF localization, place recognition, and NVS. 

As shown in \cref{tab:uav_datasets}, UAVScenes is the only dataset that simultaneously offers 6-DoF poses as well as frame-wise image and LiDAR point cloud annotations. By providing both image and LiDAR annotations with precise pose alignment, UAVScenes has the potential to significantly advance research in multi-modal UAV perception.

\subsection{Other Annotated Multi-Modal Datasets}

Besides UAV research, there are also annotated multi-modal datasets in other domains. In autonomous driving, widely used examples include KITTI~\cite{geiger2013kitti}, KITTI-360~\cite{liao2022kitti360}, nuScenes~\cite{caesar2020nuscenes}, Waymo-Open~\cite{sun2020scalability_waymo}, and K-Radar~\cite{paek2022kradar}. In robotics, popular multi-modal datasets include WildScenes~\cite{vidanapathirana2024wildscenes}, 2D-3D-Semantic~\cite{armeni2017joint2d3d}, RELLIS-3D~\cite{jiang2021rellis}, EnvoDat~\cite{nwankwo2024envodat}, and Great Outdoors~\cite{jiang2025go_greatoutdoors}. For indoor settings, EmbodiedScan~\cite{wang2024embodiedscan}, ScanNet~\cite{dai2017scannet}, ScanNet++~\cite{yeshwanth2023scannet++}, and NYU Depth v2~\cite{silberman2012indoor_nyudepthv2} are widely used.  
However, these datasets are not collected from UAV perspectives, limiting their suitability for evaluating and benchmarking various UAV tasks. To bridge this gap, UAVScenes is designed to provide a comprehensive benchmark tailored for UAV-based research.


\section{The UAVScenes Dataset}
UAVScenes builds on the MARS-LVIG \cite{li2024mars_lvig} dataset. Among existing multi-modal UAV datasets \cite{thalagala2024munfrl,xiong2024gauu_gauuscenev2,zhu2023graco,dhrafani2024firestereo}, MARS-LVIG stands out for its extensive sequential data gathered in diverse, large-scale environments—towns, valleys, airports, and islands—all traversed multiple times. This makes it ideal for benchmarking a variety of perception tasks and the most suitable foundation for our new dataset.

Although MARS-LVIG provides rich data, it primarily targets SLAM, offering only 4-DoF poses and reconstructed 3D point-cloud maps. These constraints limit its applicability to tasks requiring semantic annotations and 6-DoF poses.

To address these gaps, we extend MARS-LVIG with comprehensive camera and LiDAR semantic annotations and reconstruct 6-DoF poses with aligned 3D maps. Additionally, we benchmark six tasks and compare leading SOTA methods.

\subsection{Choices for 3D Reconstruction}

The MARS-LVIG dataset provides only 4-DoF poses using RTK, which includes a 3-DoF location and a yaw angle. As a result, MARS-LVIG is suitable solely for 4-DoF localization benchmarking as it lacks the necessary 6-DoF poses required for evaluating more fine-grained localization and reconstruction tasks, such as 6-DoF localization and NVS.

Initially, we attempt to use SOTA open-source LiDAR-visual-inertial (LVI) SLAM methods, including FAST-LIVO \cite{zheng2022fast_livo} and R3LIVE \cite{lin2022r3live}. However, ground-facing flight causes LiDAR degeneration \cite{li2024mars_lvig}, leading to unsatisfactory reconstruction results (e.g., missing too many poses, producing distorted 3D maps, and failing in reconstruction).

\begin{figure}[!htb]
\centering
\includegraphics[width=0.8\linewidth]{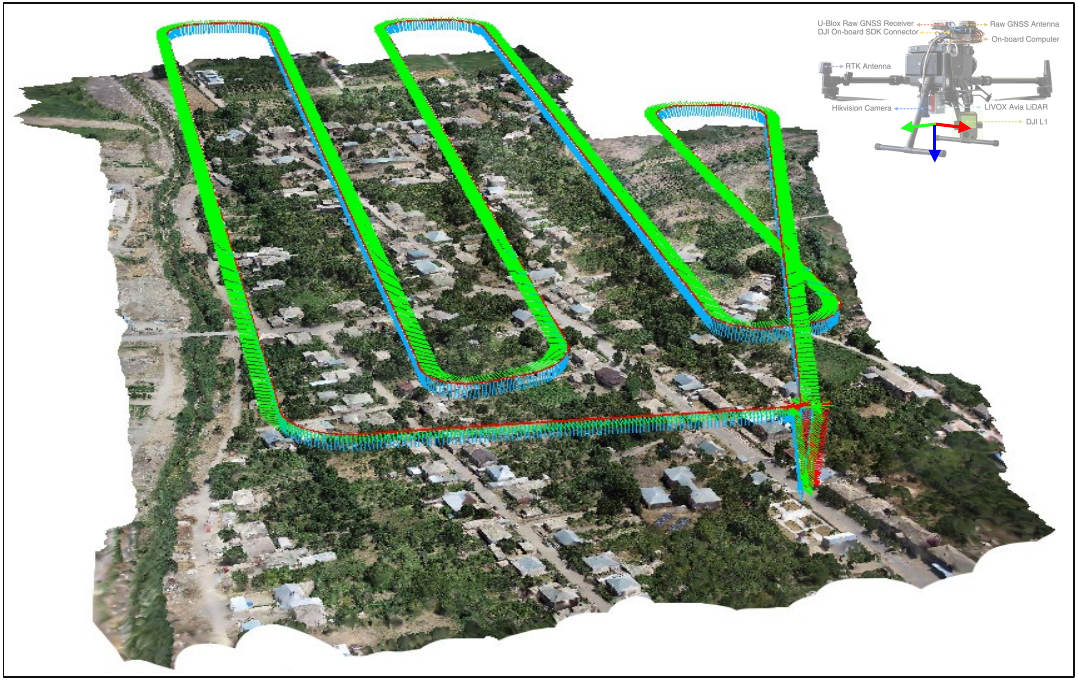}
\vspace{-8pt}
\caption{Reconstructed 3D maps and 6-DoF poses using Terra. Poses are downsampled for better visualization.}
\label{fig:3d_map}
\vspace{-8pt}
\end{figure}

\begin{figure}[!htb]
\centering
\includegraphics[width=0.9\linewidth]{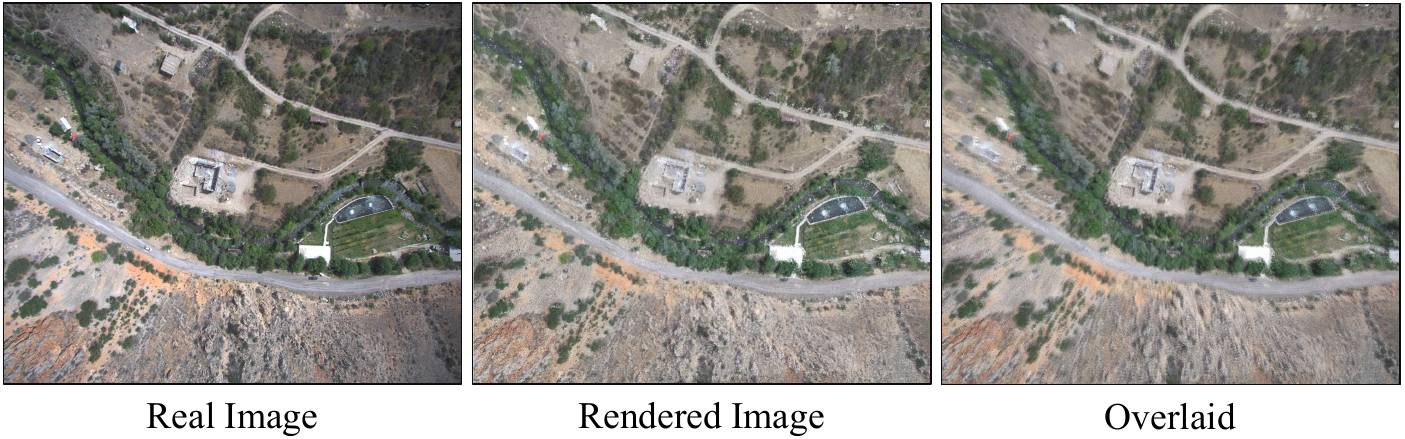}
\vspace{-8pt}
\caption{Visualization of the image rendered from the reconstructed 3D map and the 6-DoF camera pose. The rendered image closely aligns with the original image when overlaid. }
\label{fig:render_real_3d}
\vspace{-8pt}
\end{figure}

As an alternative, we use structure-from-motion (SfM) solutions to reconstruct the 6-DoF poses along with the corresponding 3D maps. We have tried various SfM solutions, including COLMAP \cite{schoenberger2016mvs_colmap}, RealityCapture\footnote{\url{https://www.capturingreality.com/}}, Metashape\footnote{\url{https://www.agisoft.com/}}, and DJI Terra\footnote{\url{https://enterprise.dji.com/dji-terra}}.
Among them, Terra, which can accept global navigation satellite system (GNSS) coordinates as the pose initializations and is specially designed for UAV scenes, provides relatively better reconstruction results. As shown in \cref{fig:3d_map,fig:render_real_3d}, the rendered image aligns well with the real captured images using the reconstructed 3D maps and 6-DoF poses.

\subsection{Frame-Wise Image Semantic Annotations}\label{subsec:2d_anno}
\subsubsection{Static Class Annotations.}
The MARS-LVIG dataset consists of multiple sensor data sequences. We need to ensure that annotations are consistent across consecutive frames. Following the annotation methodology used in SemanticKITTI \cite{behley2019semantickitti}, we divide the entire MARS-LVIG dataset into 8 distinct splits based on environmental and illumination conditions. Each split contains 1-3 sequences collected continuously within the same day, ensuring minimal scene changes except for dynamic objects. Details can be found in the supplement.

We apply Terra SfM to each split, resulting in 8 reconstructed 3D maps and their corresponding poses. The reconstruction for each split usually takes 3-10 hours\footnote{Hardware: i9-13900K + RTX 4090*2.}.
For each 3D map, we conduct manually annotating for 16 static scene classes. These annotated 3D maps are then rendered onto the corresponding camera views to produce annotated 2D semantic masks as shown in \cref{fig:2d_process}.

To ensure the quality of the rendered semantic annotations, we manually check for consistency and correct any unsatisfactory annotations. This process ensures that the image semantic annotations are both sequentially consistent and of high quality.

\subsubsection{Dynamic Class Annotations.}
\begin{figure}[!htb]
\centering
\includegraphics[width=0.99\linewidth]{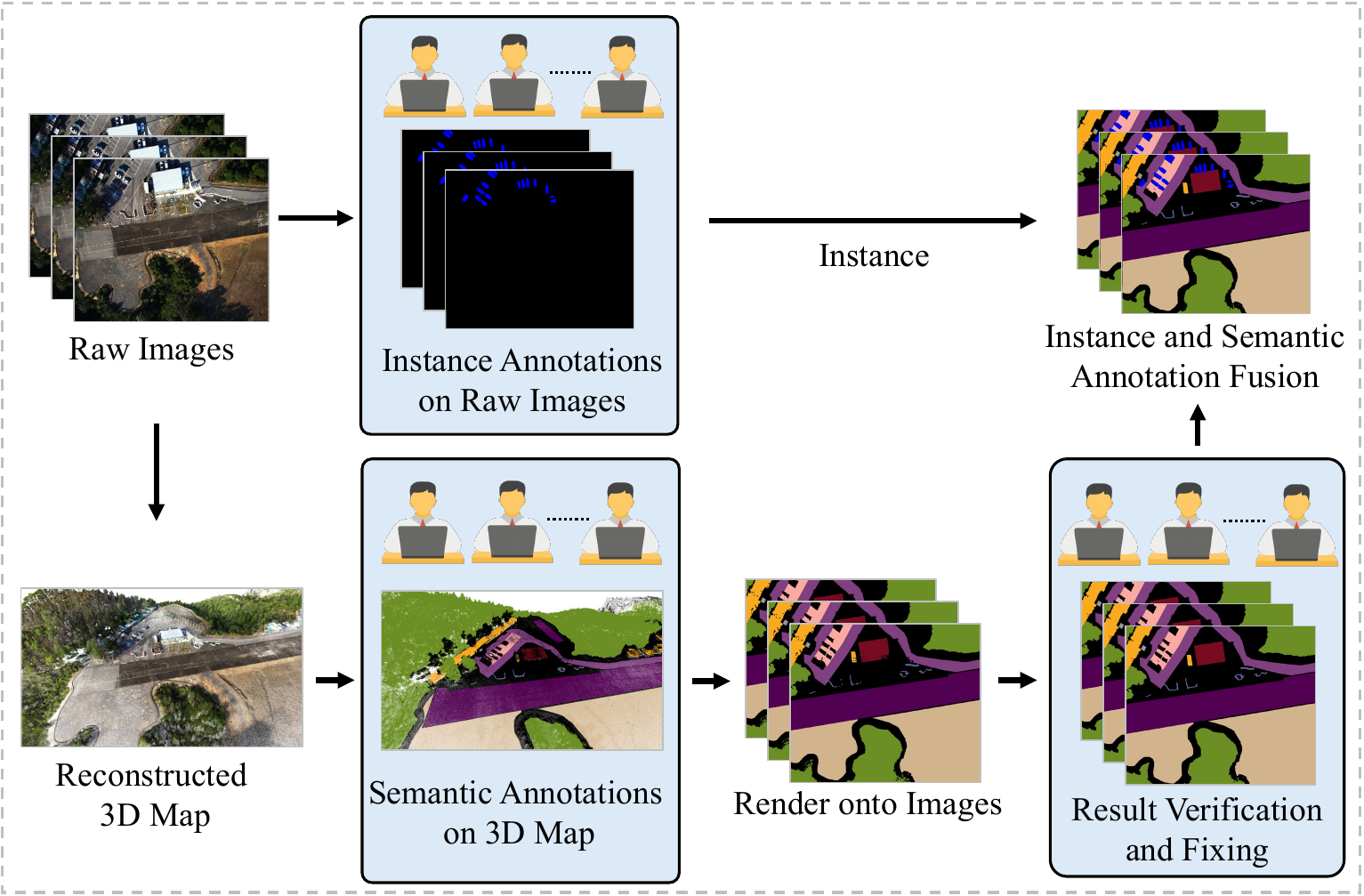}
\vspace{-8pt}
\caption{The 2D image annotating pipeline. Manual annotating is conducted at 3D map annotations, instance annotations, and fixing stages. }
\label{fig:2d_process}
\vspace{-8pt}
\end{figure}

Since the rendered static scene masks do not account for dynamic objects, we manually annotate instance-wise labels for 2 dynamic object classes (sedan and truck) in each image as shown in \cref{fig:2d_process}. 
As MARS-LVIG is sequentially captured, manual annotating can be partially accelerated by auto-labeling tracking (tracking is always unstable), followed by human verification and fixing. We use X-AnyLabeling\footnote{\url{https://github.com/CVHub520/X-AnyLabeling}} to achieve tracking. 
In total, we have manually annotated over 280k dynamic instances in the dataset (see the statistics in \cref{tab:instance_stats}). The 2D static semantic annotations and 2D dynamic instance annotations are then combined to produce the final 2D annotations for each image.

\begin{figure}[!htb]
\centering
\includegraphics[width=0.99\linewidth]{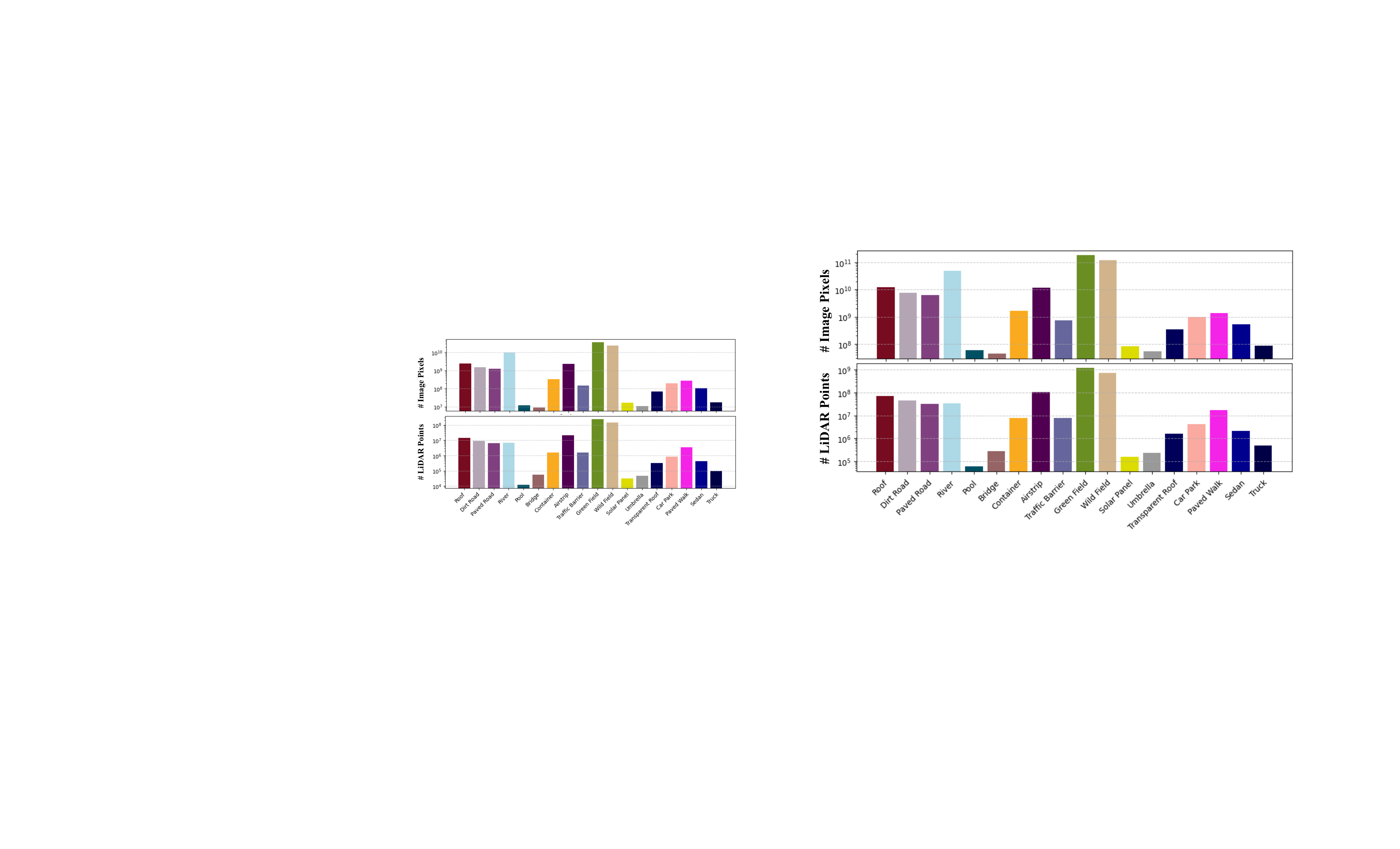}
\vspace{-8pt}
\caption{The annotation class distribution visualization. The background class is ignored. Above: image pixel annotations. Below: LiDAR point cloud annotations. 
}
\label{fig:statistics}
\vspace{-8pt}
\end{figure}

This annotation process results in 120k annotated images with 19 classes, including 16 static classes, 2 dynamic classes, and 1 background class. The overall class distribution is shown in \cref{fig:statistics}.

\begin{table}[!htb]
\vspace{-4pt}
\scriptsize
\centering
\begin{tabular}{l|cc}
\toprule
& Sedan & Truck \\
\midrule
Avg. BBox Height (pixel)      & 72 & 106\\
Avg. BBox Width (pixel)      & 68 & 106\\
Avg. BBox Area (pixel)       & 5001 & 12411\\
Avg. Polygon Area (pixel)    & 3210 & 6873\\
Avg. Occupancy Ratio  & 67\% & 62\%\\
\midrule
\#Instances & 270k & 14k\\ 
\bottomrule
\end{tabular}
\vspace{-8pt}
\caption{Image instance statistics for dynamic object classes.}
\vspace{-8pt}
\label{tab:instance_stats}
\end{table}


\subsection{Frame-Wise LiDAR Point Cloud Annotations}

The LiDAR is a crucial sensor for multi-modal perception. It casts laser beams to capture the spatial information\footnote{Some LiDARs also provide other information, e.g., intensity.} of the environment. 

In the MARS-LVIG dataset, there are two distinct LiDAR sensors: the DJI-L1 and the Livox-Avia. Due to manufacturer-imposed encryption on the DJI-L1's output data, access to its raw point clouds is restricted, limiting its utility for open research. Therefore, our dataset annotations are focused exclusively on data captured by the open-source Livox-Avia LiDAR, which enables unrestricted access to high-quality point cloud data, facilitating broader applicability and reproducibility in academic and industrial research.

\begin{figure}[t]
\centering
\includegraphics[width=0.99\linewidth]{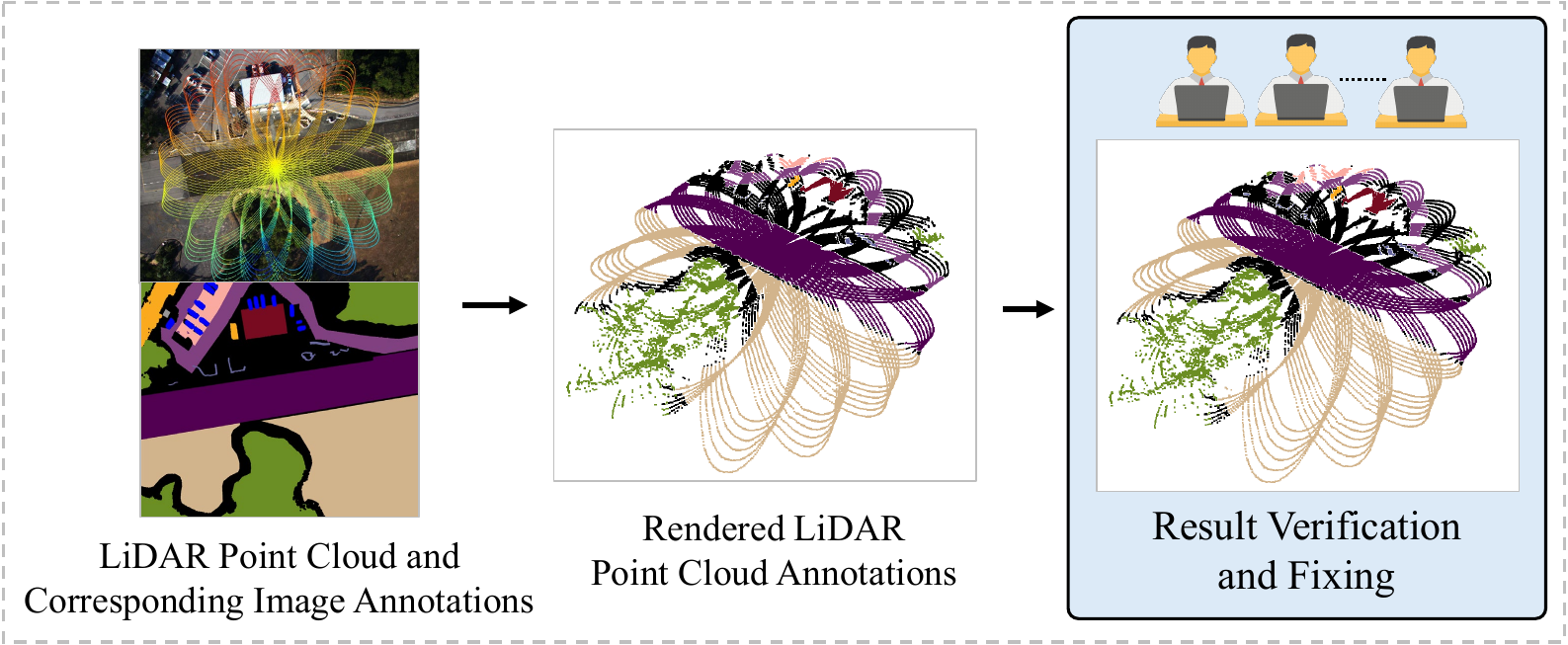}
\vspace{-8pt}
\caption{The LiDAR point cloud annotating pipeline. Manual annotating is conducted at fixing stages.}
\label{fig:3d_process}
\vspace{-8pt}
\end{figure}

The MARS-LVIG dataset is acquired using a hardware-synchronized camera-LiDAR suite, which incorporates a calibrated camera and LiDAR sensors. By leveraging this calibration, we project image annotations onto the corresponding LiDAR point clouds. Following the procedure described in \cref{subsec:2d_anno}, we conduct thorough consistency checks, manually correcting any unsatisfactory annotations within the LiDAR point clouds to ensure high fidelity between each camera-LiDAR frame pair. This workflow is illustrated in \cref{fig:3d_process}. The class distribution for LiDAR point cloud annotations is shown in \cref{fig:statistics}.

\section{Benchmark Experiments}
In this section, we establish benchmarks on various perception tasks using the proposed UAVScenes dataset. The existing benchmark tasks include frame-wise image and LiDAR semantic segmentation, place recognition, depth estimation, 6-DoF localization, and NVS. 


\subsection{Image Semantic Segmentation}

Image semantic segmentation is a fundamental task in computer vision and is essential for evaluating the performance of vision models. It involves predicting the class label for each pixel in an input image. We consider several backbone architectures, including ResNet \cite{he2016resnet}, ConvNext \cite{liu2022convnext}, ConvNextV2 \cite{woo2023convnextv2}, ViT \cite{dosovitskiy2020image_vit}, MambaOut \cite{yu2024mambaout}, and DeiT3 \cite{touvron2022deit3_deitiii}. We use UperNet \cite{xiao2018unified_upernet} as the segmentation head, which is widely used in semantic segmentation evaluations. All models are based on the TIMM package\footnote{\url{https://github.com/huggingface/pytorch-image-models}}.

\begin{figure}
\centering
\includegraphics[width=0.9\linewidth]{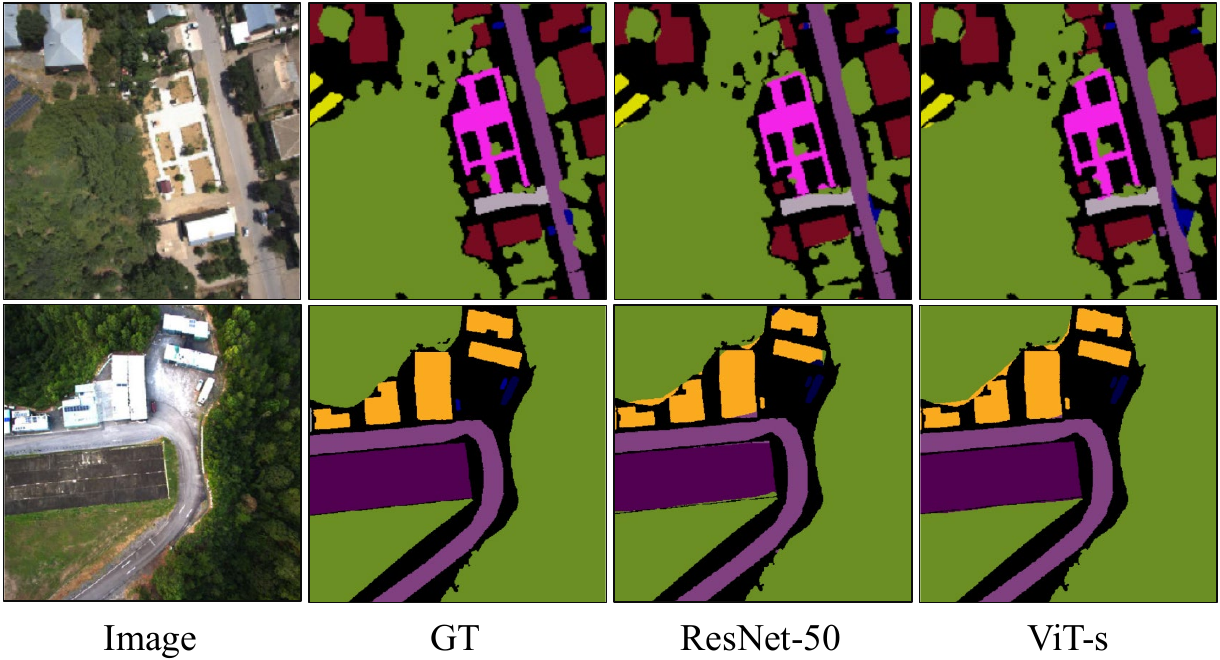}
\vspace{-8pt}
\caption{Visualization of Image semantic segmentation results.}
\vspace{-8pt}
\label{fig:viz_2dseg}
\end{figure}

As shown in \cref{stab:2dseg_3gseg_results}, among these backbones, DeiT3 \cite{dosovitskiy2020image_vit} achieves the best performance. Moreover, the Transformer-based models generally outperform convolutional neural networks (CNNs), which demonstrate the effectiveness of Transformers. We visualize some examples in \cref{fig:viz_2dseg}.


\begin{table*}[htbp]
\centering
\scriptsize
\setlength{\tabcolsep}{0.05cm} 
\begin{tabular}{r|c|l|c|cccccccccccccccccc}
\toprule
\multirow{4}{*}{\#Params} & \multirow{4}{*}{Arch.} & \multirow{4}{*}{Model} & \multirow{4}{*}{mIoU$\uparrow$} & \multicolumn{18}{c}{Per Class IoU$\uparrow$}\\
\cline{5-22}
&&&& \makecell{Roof} & \makecell{Dirt \\ Road} & \makecell{Paved \\ Road} & \makecell{River} & \makecell{Pool} & \makecell{Bridge} & \makecell{Conta.} & \makecell{Airstrip} & \makecell{Traffic \\ Barrier} & \makecell{Green \\ Field} & \makecell{Wild \\ Field} & \makecell{Solar \\ Panel} & \makecell{Umbre.} & \makecell{Transp. \\ Roof} & \makecell{Car \\ Park} & \makecell{Paved \\ Walk} & \makecell{Sedan} & \makecell{Truck} \\
& &  & & \cellcolor{roof} & \cellcolor{dirt} & \cellcolor{pavedmotor} & \cellcolor{river} & \cellcolor{pool} & \cellcolor{bridge} & \cellcolor{container} & \cellcolor{airstrip} & \cellcolor{traffic} & \cellcolor{greenfield} & \cellcolor{wildfield} & \cellcolor{solar} & \cellcolor{umbrella} & \cellcolor{transparent} & \cellcolor{carpark} & \cellcolor{pavedwalk} & \cellcolor{sedan} & \cellcolor{truck} \\
\midrule
\multicolumn{22}{c}{Image Semantic Segmentation} \\
\midrule
21M & CNN & ResNet-34 \cite{he2016resnet} & 59.9 & 74.3 & 53.9 & 77.4 & 91.6 & 25.4 & 21.4 & 69.9 & 88.1 & 53.8 & 89.8 & 87.4 & 74.7 & 2.1 & 64.2 & 52.0 & 89.1 & 22.3 & 41.5 \\
25M & CNN & ResNet-50 \cite{he2016resnet} & 61.3 & 76.0 & 52.6 & 77.8 & 88.4 & 19.3 & 30.9 & 69.3 & 91.9 & 49.4 & 90.4 & 88.6 & 77.9 & 8.5 & 65.8 & 51.9 & 94.5 & 20.1 & 49.8 \\
44M & CNN & ResNet-101 \cite{he2016resnet} & 60.7 & 77.0 & 53.9 & 78.1 & 75.9 & 29.2 & 33.8 & 70.3 & 92.6 & 54.0 & 91.0 & 80.9 & 80.4 & 8.3 & 66.2 & 50.3 & 91.1 & 16.5 & 43.3 \\
28M & CNN & ConvNext-t \cite{liu2022convnext}& 55.3 & 72.4 & 46.4 & 71.1 & 84.4 & 18.1 & 20.7 & 64.2 & 82.8 & 46.2 & 91.2 & 84.4 & 79.1 & 1.1 & 61.4 & 39.9 & 81.1 & 10.5 & 39.8 \\
28M & CNN & ConvNextV2-t \cite{woo2023convnextv2} & 53.1 & 70.8 & 44.6 & 67.4 & 88.0 & 21.0 & 18.0 & 55.9 & 80.1 & 43.6 & 90.7 & 86.6 & 72.3 & 5.8 & 56.9 & 27.6 & 80.0 & 7.8 & 38.5 \\
48M & CNN & MambaOut-s \cite{yu2024mambaout} & 51.8 & 65.2 & 46.6 & 69.9 & 56.5 & 25.4 & 19.0 & 58.1 & 78.0 & 36.5 & 82.3 & 82.7 & 76.2 & 2.6 & 57.8 & 44.1 & 79.9 & 12.0 & 39.2 \\
26M & CNN & MambaOut-t \cite{yu2024mambaout} & 50.0 & 59.0 & 43.1 & 63.5 & 65.6 & 19.1 & 20.0 & 55.9 & 74.0 & 34.4 & 80.0 & 81.0 & 76.1 & 1.3 & 57.8 & 40.5 & 80.0 & 12.0 & 37.1 \\
5M & Transf. & ViT-t \cite{dosovitskiy2020image_vit} &  62.8 & 74.3 & 58.8 & 76.7 & 90.9 & 41.2 & 52.8 & 51.3 & 80.4 & 39.4 & 93.6 & 90.3 & 88.9 & 19.3 & 76.4 & 62.3 & 86.4 & 26.5 & 20.3 \\
22M & Transf. & ViT-s \cite{dosovitskiy2020image_vit} & 63.9 & 75.0 & 61.2 & 77.4 & 88.7 & 49.0 & 54.9 & 56.5 & 86.5 & 51.4 & 94.3 & 90.0 & 89.5 & 11.2 & 80.5 & 52.4 & 89.7 & 20.2 & 21.9 \\
22M & Transf. & DeiT3-s \cite{touvron2022deit3_deitiii} & 67.6 & 76.0 & 67.0 & 81.1 & 91.0 & 58.1 & 57.8 & 62.7 & 88.0 & 41.6 & 91.1 & 91.7 & 90.0 & 24.1 & 82.9 & 63.2 & 93.0 & 28.1 & 30.0 \\
38M & Transf. & DeiT3-m \cite{touvron2022deit3_deitiii} & \first{68.3} & 77.6 & 66.2 & 79.3 & 92.2 & 52.3 & 56.6 & 58.9 & 88.6 & 53.2 & 93.6 & 92.4 & 90.1 & 30.9 & 83.5 & 60.4 & 93.7 & 27.3 & 32.5 \\
\midrule
\multicolumn{22}{c}{Frame-wise LiDAR Semantic Segmentation} \\
\midrule
38M & - & MinkUNet \cite{choy20194d_mink} & 32.7 & 74.5 & 43.4 & 57.6 & 61.3 & 0.0 & 10.3 & 14.4 & 47.3 & 32.3 & 86.2 & 81.8 & 2.3 & 1.4 & 31.1 & 9.9 & 18.3 & 13.4 & 3.1 \\
39M & - & SPUNet \cite{spconv2022} & \first{34.4} & 73.9 & 38.1 & 56.3 & 37.0 & 0.0 & 15.1 & 38.5 & 65.4 & 38.8 & 85.7 & 78.1 & 0.0 & 0.0 & 23.0 & 8.4 & 47.2 & 13.0 & 0.0 \\
11M & - & PTv2 \cite{wu2022point_ptv2} & {33.2} & 71.7 & 38.4 & 32.7 & 38.2 & 0.0 & 8.6 & 47.9 & 34.2 & 50.1 & 75.1 & 55.0 & 3.0 & 53.8 & 41.3 & 2.0 & 0.1 & 27.1 & 18.4 \\
\bottomrule
\end{tabular}
\vspace{-8pt}
\caption{Semantic segmentation results with mIoU (\%) and class IoU (\%). Above: Camera-Based. Below: LiDAR-Based.}
\vspace{-8pt}
\label{stab:2dseg_3gseg_results}
\end{table*}

\subsection{Frame-wise LiDAR Semantic Segmentation} 

Frame-wise LiDAR point cloud semantic segmentation is a crucial task for 3D scene understanding, involving the prediction of class labels for each point in the LiDAR-generated point cloud. For this task, we run and evaluate three baseline models: MinkUNet~\cite{choy20194d_mink}, SPUNet \cite{spconv2022}, and PTv2 \cite{wu2022point_ptv2}. All models are based on the Pointcept package\footnote{\url{https://github.com/Pointcept/Pointcept}}.

\begin{figure}
\centering
\vspace{-3pt}
\includegraphics[width=0.9\linewidth]{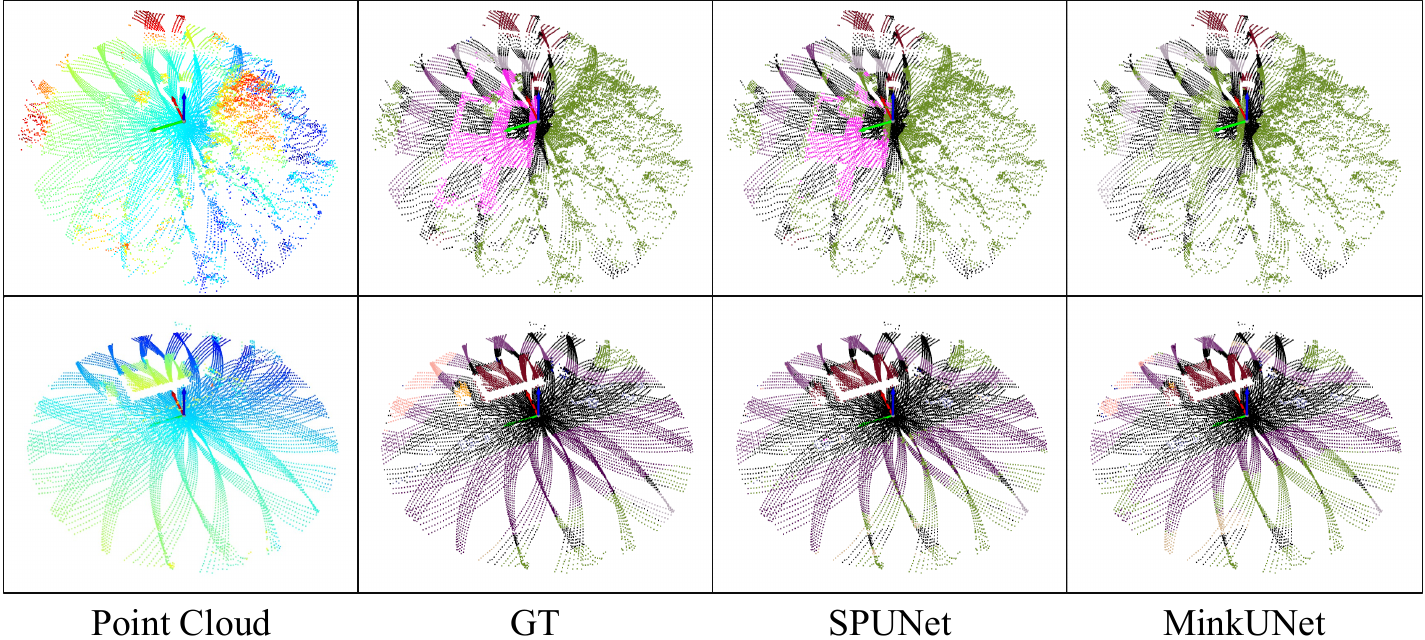}
\vspace{-8pt}
\caption{Visualization of LiDAR semantic segmentation results.}
\vspace{-8pt}
\label{fig:viz_3dseg}
\end{figure}

As shown in \cref{stab:2dseg_3gseg_results}, the three networks show comparable performance. PTv2 with the least number of parameters (11M) can surpass MinkUNet (38M), demonstrating the network architecture effectiveness. In addition, the pool class would be a challenging class for LiDAR semantic segmentation as all three networks show 0 class IoU. This would be attributed to the low quantity of annotated LiDAR point clouds as in \cref{fig:statistics}. We visualize some LiDAR semantic segmentation examples in \cref{fig:viz_3dseg}.


\subsection{Place Recognition} 

Place recognition treats localization as a retrieval problem \cite{arandjelovic2016netvlad}. In this approach, the place is recognized by matching a query image to a database of images, with the location of the top-matched database image being regarded as the query's location. In this task, we compare camera-based, LiDAR-based, and fusion-based place recognition methods. The image-based models include GeM \cite{radenovic2018gempooling}, RRM \cite{kordopatis2021leveraging_rrm}, ConvAP \cite{ali2022convap}, MixVPR \cite{ali2023mixvpr}, AnyLoc \cite{keetha2023anyloc}, and SALAD \cite{izquierdo2024salad}. The LiDAR-based models include MinkLoc3D \cite{komorowski2021minkloc3d}, MinkLoc3D V2 \cite{komorowski2022minkloc3dv2}, and BEVPlace \cite{luo2023bevplace}. The fusion-based models include MinkLoc++ \cite{komorowski2021minkloc++}, AdaFusion \cite{lai2022adafusion}, LCPR \cite{zhou2024lcpr}, and UMF \cite{umf2024}.

In \cref{tab:place_recognition}, we compare the recall performance of different models. Generally, fusion-based place recognition models outperform their single-modal counterparts, demonstrating the effectiveness of multi-modal fusion. For camera-based place recognition models, the inclusion of strong foundation backbones like DINO V2 \cite{oquab2023dinov2} brings significant improvements. Additionally, the projection model BEVPlace performs worse than the voxel model MinkLoc3D, indicating that BEV LiDAR projection may not be a suitable format for place recognition under UAV perspectives.

\begin{table}[t]
\scriptsize
\centering
\setlength{\tabcolsep}{0.05cm} 
\begin{tabular}{l | c | c c c }
\toprule
Model & Modality &  Recall@1$\uparrow$ & Recall@5$\uparrow$  & Recall@10$\uparrow$  \\
\midrule
GeM \cite{radenovic2018gempooling}    & C & 42.1 & 55.8 & 62.0 \\ 
RRM \cite{kordopatis2021leveraging_rrm}    & C & 41.7 & 53.4 & 61.1 \\ 
ConvAP \cite{ali2022convap}  & C & 41.1 & 54.8 & 63.0 \\ 
MixVPR \cite{ali2023mixvpr} & C & 34.0 & 53.0 & 61.6 \\ 
\midrule
AnyLoc \cite{keetha2023anyloc} (DINO V2-s \cite{oquab2023dinov2})  & C & 58.5 & 74.4 & 79.1 \\
SALAD \cite{izquierdo2024salad} (DINO V2-s \cite{oquab2023dinov2})   & C & \first{67.1} & \first{76.4} & \first{79.8} \\
\midrule
MinkLoc3D \cite{komorowski2021minkloc3d} & L    &  41.9 & 60.0 & 66.7 \\
MinkLoc3D V2 \cite{komorowski2022minkloc3dv2} & L &  42.8 & 61.5 & 67.3 \\
BEVPlace \cite{luo2023bevplace} & L     &  32.6 & 54.6 & 64.2 \\
\midrule
MinkLoc++ \cite{komorowski2021minkloc++}& C+L & 47.1 & 63.5 & 69.0 \\ 
AdaFusion \cite{lai2022adafusion} & C+L & 46.3 & 63.4 & 70.2 \\ 
LCPR \cite{zhou2024lcpr}     & C+L & 42.3 & 62.3 & 68.8 \\ 
UMF \cite{umf2024}      & C+L & 40.1 & 53.9 & 61.0 \\ 
\bottomrule
\end{tabular}
\vspace{-8pt}
\caption{Place recognition performance with Recall@$K$ ($K=1,5,10$) (\%). AnyLoc and SALAD use the visual foundation backbone DINO V2\cite{oquab2023dinov2}, while other models use ResNet-18 as the 2D backbone.}
\label{tab:place_recognition}
\vspace{-8pt}
\end{table}


\subsection{Novel View Synthesis} 

NVS generates new perspectives of a scene from limited image viewpoints, paving the way for realistic and efficient 3D scene generation that captures intricate lighting, textures, and geometric details. NVS is largely driven by neural radiance fields (NeRFs) \cite{mildenhall2021nerf}, which model 3D scenes as continuous functions with differentiable rendering, and 3D Gaussians (GS) \cite{kerbl:hal-04088161}, which represent scenes as learnable 3D Gaussians for rasterized rendering.
To evaluate NVS, we introduce NeRFs-based and 3D GS-based baselines: Instant-NGP \cite{mueller2022instant}, 3DGS \cite{kerbl:hal-04088161}, GaussianPro \cite{GaussianPro}, DCGaussian \cite{wang2024dc}, and Pixel-GS \cite{zhang2024pixelgs}.

\begin{figure}[t]
\centering
\includegraphics[width=0.99\linewidth]{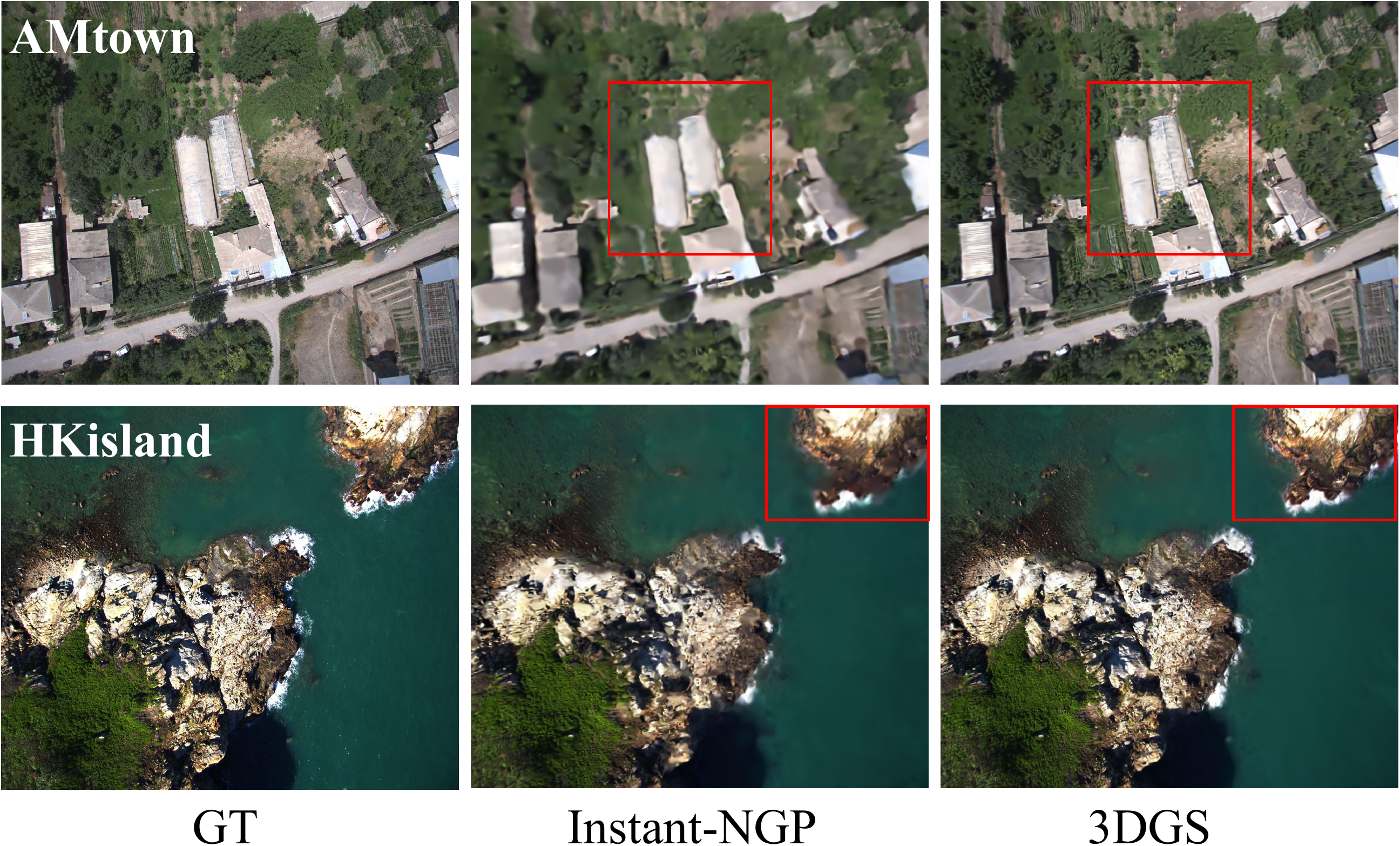}
\vspace{-8pt}
\caption{Qualitative evaluation of NVS. The areas outlined in red highlight regions with significant rendering discrepancies.}
\label{fig_nvs}
\vspace{-8pt}
\end{figure}

The quantitative results and qualitative visualizations on UAVScenes are presented in \cref{tab_tab_ss} and \cref{fig_nvs}, respectively.
All 3D GS-based methods use the raw point cloud provided by the dataset as the initialization.
The NeRF method Instant-NGP performs poorly on large-scale aerial images.
The 3D GS methods, 3DGS and Pixel-GS, achieve better rendering performance than others.
However, in certain areas, such as adjacent buildings and repetitive forest scenes, the performance of NVS still requires improvement, as highlighted by red boxes.

\begin{table*}[!htb]
\vspace{-8pt}
\scriptsize
\begin{center}
\begin{tabular}{l|ccc|ccc|ccc|ccc}
\toprule
\multicolumn{1}{l|}{\multirow{2}{*}{Model}} & \multicolumn{3}{c|}{AMtown} & \multicolumn{3}{c|}{AMvalley} & \multicolumn{3}{c|}{HKairport} & \multicolumn{3}{c}{HKisland} \\ & PSNR$\uparrow$  & SSIM$\uparrow$ & LPIPS$\downarrow$ & PSNR$\uparrow$  & SSIM$\uparrow$ & LPIPS$\downarrow$ & PSNR$\uparrow$  & SSIM$\uparrow$ & LPIPS$\downarrow$ & PSNR$\uparrow$  & SSIM$\uparrow$ & LPIPS$\downarrow$ \\
\midrule  
Instant-NGP \cite{mueller2022instant} & 20.09 & 0.451 & 0.674 & 23.20 & 0.497 & 0.622 & 19.39 & 0.442 & 0.604 & 16.81 & 0.508 & 0.556 \\ 
3DGS \cite{kerbl:hal-04088161} & \textbf{22.95} & \textbf{0.547} & 0.551  & \textbf{25.12} & \textbf{0.576} & 0.514 & 20.92 & \textbf{0.523} & 0.481 & \textbf{17.85} & \textbf{0.553} & 0.494 \\ 
GaussianPro \cite{GaussianPro} & 22.83 & 0.546 & \textbf{0.549} & 25.09 & \textbf{0.576} & \textbf{0.511} & 20.86 & 0.521 & \textbf{0.478} & 17.82 & 0.552 & \textbf{0.491} \\ 
DCGaussian \cite{wang2024dc} & 22.92  & 0.537 & 0.556 & 25.10 & 0.573 & 0.516 & 20.81 & 0.510 & 0.495 & 17.77 & 0.546 & 0.503 \\ 
Pixel-GS \cite{zhang2024pixelgs} & \textbf{22.95} & \textbf{0.547} & 0.551  & 25.11 & \textbf{0.576} & 0.514 & \textbf{20.93} & \textbf{0.523} & 0.481 & 17.84 & \textbf{0.553} & 0.494 \\ 
\bottomrule 
\end{tabular}
\vspace{-8pt}
\caption{Quantitative evaluation of NVS. The evaluated metrics include PSNR, SSIM, and LPIPS.}
\label{tab_tab_ss}
\end{center}
\vspace{-8pt}
\end{table*}

\subsection{6-DoF Visual Localization} 

6-DoF visual localization is a fundamental task in computer vision, essential for applications like robotics and augmented reality. 
Its goal is to estimate the 6-DoF pose of a query image within a pre-existing environment map.
Currently, absolute pose regression (APR) and scene coordinate regression (SCR) have made significant strides in localization. 
APR estimates the 6-DoF pose of an input image through direct regression, enabling an end-to-end, highly efficient localization process.
SCR localizes by regressing the 3D coordinates of 2D image pixels rather than directly estimating the camera pose, enabling training via re-projection error. 
The camera pose is then determined through 2D-3D correspondences.
This paper conducts experiments using modern APR baselines, including PoseNet \cite{posenet}, AtLoc \cite{wang2020atloc}, and RobustLoc \cite{wang2023robustloc}, as well as SCR baselines such as ACE \cite{brachmann2023ace}, GLACE \cite{GLACE2024CVPR}, and FocusTune \cite{10483746}.

\begin{figure}[!htb]
\centering
\includegraphics[width=0.9\linewidth]{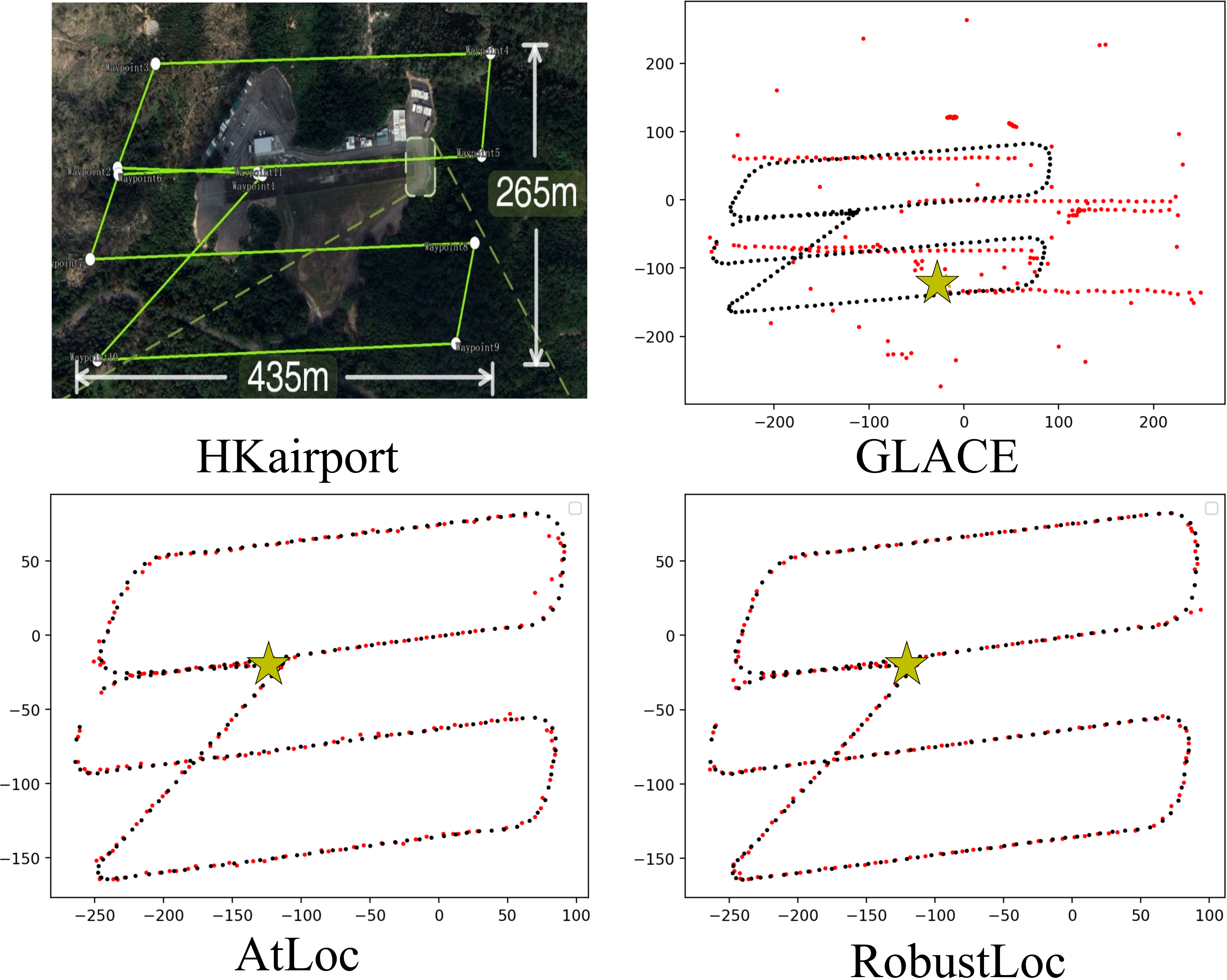}
\vspace{-8pt}
\caption{Visualization of 6-DoF localization. The ground truth and prediction are black and red lines, respectively. The star denotes the first frame. Metrics are in meters.}
\label{fig_scr}
\vspace{-8pt}
\end{figure}

The localization errors (position and rotation) and qualitative visualizations on UAVScenes are presented in \cref{tab_tab_scr} and \cref{fig_scr}, respectively.  
All APR-based methods demonstrate strong performance, with RobustLoc achieving the best results, significantly outperforming others.  
SCR-based methods use a frozen pre-trained encoder for faster training, but its ground-urban-scene pretraining limits performance on UAV-view images, leading to higher localization errors.


\begin{table}[t]
\vspace{-8pt}
\scriptsize
\setlength{\tabcolsep}{0.1cm} 
\begin{center}
\begin{tabular}{l|ccccc}
\toprule
Model & AMtown & AMvalley & HKairport & HKisland & Average\\ 
\midrule  
ACE \cite{brachmann2023ace} & 180.8 / 1.5 & 145.8 / 0.6 & 83.0 / 0.6 & 121.9 / 0.8 & 132.9 / 0.9 \\ 
FocusTune \cite{10483746} & 188.9 / 1.0 & 156.0 / 0.6 & 72.1 / 0.6 & 113.7 / 0.9 & 132.7 / 0.8 \\ 
GLACE \cite{GLACE2024CVPR} & 90.9 / 0.6 & 90.1 / 0.4 & 58.3 / 0.5 & 102.8 / 0.7 & 85.5 / 0.6 \\ 
PoseNet \cite{posenet} & 43.1 / 0.2 & 22.7 / 0.2 & 14.9 / 0.2 & 22.5 / 0.1 & 25.8 / 0.2 \\ 
AtLoc \cite{wang2020atloc} & 12.7 / 0.2 & \first{9.0} / \first{0.1} & 6.0 / \first{0.1} & 6.7 / \first{0.1} & 8.6 / \first{0.1} \\ 
RobustLoc \cite{wang2023robustloc} & \first{5.9 / 0.1} & 9.8 / \first{0.1} & \first{4.9 / 0.1} & \first{3.6 / 0.1} & \first{6.1 / 0.1} \\ 
\bottomrule 
\end{tabular}
\vspace{-8pt}
\caption{Quantitative evaluation of visual localization on the UAVScenes dataset. We report median position error (m) and median rotation error (degree). }
\label{tab_tab_scr}
\end{center}
\vspace{-8pt}
\end{table}


\subsection{Depth Estimation}
Depth estimation involves predicting pixel-wise depth values from input images, bridging the gap between 2D imagery and 3D spatial understanding. This task is particularly valuable for evaluating camera-only perception systems that require real-time or lightweight operation (e.g., lightweight cameras-only UAVs).  
Since the MARS-LVIG dataset does not provide such evaluation (though its calibrated camera-LiDAR suite can support), we add this task to create a more comprehensive benchmark.  
In this section, we evaluate zero-shot depth estimation models to assess their generalization capabilities in UAV aerial views. We consider both single-step models and diffusion-based multi-step models. The single-step models include ZoeDepth~\cite{bhat2023zoedepth}, Depth Anything~\cite{bhat2023zoedepth}, Depth Anything V2~\cite{yang2024depthanythingv2}, Metric3D~\cite{yin2023metric3d}, and Metric3D V2~\cite{hu2024metric3dv2}. The diffusion models include GeoWizard~\cite{fu2025geowizard} and Marigold~\cite{ke2024repurposing_marigold}. The ground truth depth for each image is derived from the corresponding LiDAR frame.

\begin{figure}[!htb]
\centering
\includegraphics[width=0.8\linewidth]{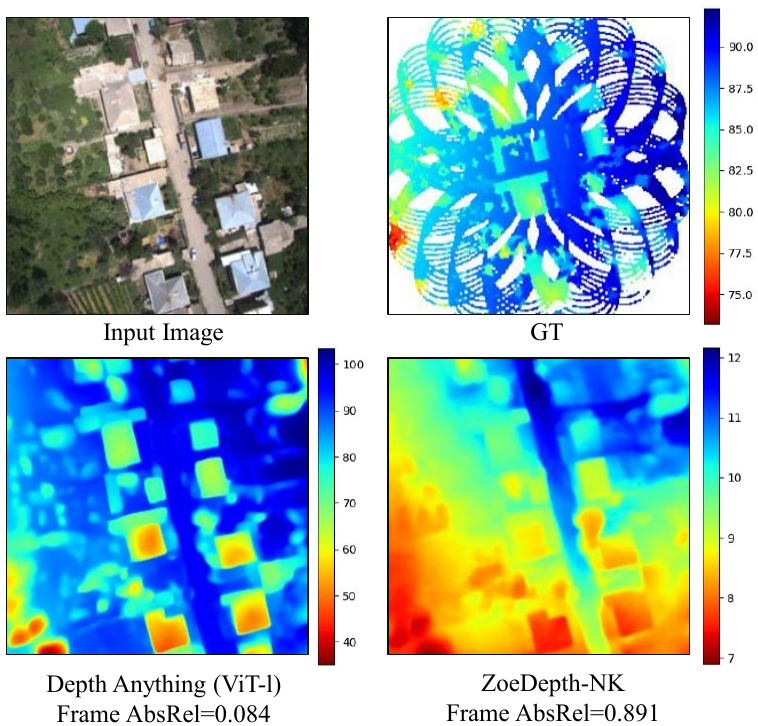}
\vspace{-8pt}
\caption{Visualization of the zero-shot depth estimation results. The depth ground truth is from the corresponding LiDAR point cloud. The color bar indicates depth values.}
\label{fig:depth_estimation}
\vspace{-8pt}
\end{figure}

As shown in \cref{tab:depth_estimation}, Metric3D V2 demonstrates the best performance in terms of absolute relative error and square relative error. However, Depth Anything V2 outperforms it in the $\delta_1$ metric. For diffusion-based models, which only support affine-invariant depth maps, the performance is relatively worse compared to their single-step counterparts. We visualize the depth predictions of the single-step models in \cref{fig:depth_estimation}.
Most zero-shot monocular depth estimation schemes lack generalization ability and accuracy in the UAV perspective, underscoring the need for advancements in this area.

\begin{table}[!htb]
\scriptsize
\centering
\begin{tabular}{l | c c c }
\toprule
Model &  AbsRel$\downarrow$ & SqRel$\downarrow$  & $\delta_1$$\uparrow$  \\
\midrule
ZoeDepth-K \cite{bhat2023zoedepth} & 0.976 & 81.752 & 0 \\
ZoeDepth-N \cite{bhat2023zoedepth} & 0.975 & 81.508 & 0 \\
ZoeDepth-NK \cite{bhat2023zoedepth} & 0.894 & 69.939 & 0 \\
Depth Anything (ViT-b) \cite{yang2024depthanything} & 0.707 & 46.102 & 0.010 \\
Depth Anything (ViT-l) \cite{yang2024depthanything} & 0.472 & 36.029 & 0.453 \\
Depth Anything V2 (ViT-b) \cite{yang2024depthanything} & 0.939 & 76.630 & \first{1.670} \\
Depth Anything V2 (ViT-l) \cite{yang2024depthanything} & 1.517 & 261.925 & 0.089 \\
Metric3D (ConvNeXt-t) \cite{yin2023metric3d} & 0.790 & 58.411 & 0.009 \\
Metric3D (ConvNeXt-l) \cite{yin2023metric3d} & 0.682 & 53.504 & 0.160 \\
Metric3D V2 (ViT-s) \cite{hu2024metric3dv2} & 0.830 & 68.084 & 0.028 \\
Metric3D V2 (ViT-l) \cite{hu2024metric3dv2} & \first{0.540} & \first{31.960} & 0.074 \\
Marigold  \cite{ke2024repurposing_marigold} & 0.994 & 84.409 & 0\\
GeoWizard  \cite{fu2025geowizard} & 0.995 & 84.485 & 0\\
\bottomrule
\end{tabular}
\vspace{-8pt}
\caption{Zero-shot depth estimation performance on the UAVScenes dataset. "l" denotes large. Above are single-step models, and below Marigold and GeoWizard are diffusion-based models that can only produce affine-invariant depth maps. Evaluation metrics follow MonoDepth2 \cite{monodepth2}.}
\label{tab:depth_estimation}
\vspace{-8pt}
\end{table}

\section{Limitation and Conclusion}
Although UAVScenes has captured large-scale environments, expanding its diversity remains crucial. Future efforts could include complex urban or downtown areas with varied streets, vehicles, high-rise buildings, and pedestrians. 


UAVScenes is a versatile multi-modal UAV dataset that provides rich semantic annotations for both 2D images and 3D LiDAR point clouds. With precisely aligned 6-DoF poses and associated 3D maps, it accommodates diverse research needs. By introducing a standardized benchmark for UAV perception tasks, UAVScenes offers consistent evaluation and comparison across multiple modalities. It thus serves as a fundamental resource for advancing UAV perception and mapping, driving progress in autonomous navigation, scene comprehension, and cross-modal learning within the UAV field.



{
    \small
    \bibliographystyle{ieeenat_fullname}
    \bibliography{ref}
}





\renewcommand{\thesection}{S\arabic{section}}
\renewcommand{\thesubsection}{S\arabic{section}.\arabic{subsection}}
\renewcommand{\thesubsubsection}{S\arabic{section}.\arabic{subsection}.\arabic{subsection}}

\renewcommand{\thefigure}{S\arabic{figure}}
\renewcommand{\thetable}{S\arabic{table}}
\renewcommand{\theequation}{S\arabic{equation}}

\section{Dataset Details}\label{ssec:dataset_details}
In this section, we provide more details about data collection and processing. Our datasets and baseline codes (also with train/test split instructions) are available at: \url{https://github.com/sijieaaa/UAVScenes}

\subsection{Data Collection Platform}
\begin{figure}[!htb]
\centering
\includegraphics[width=0.6\linewidth]{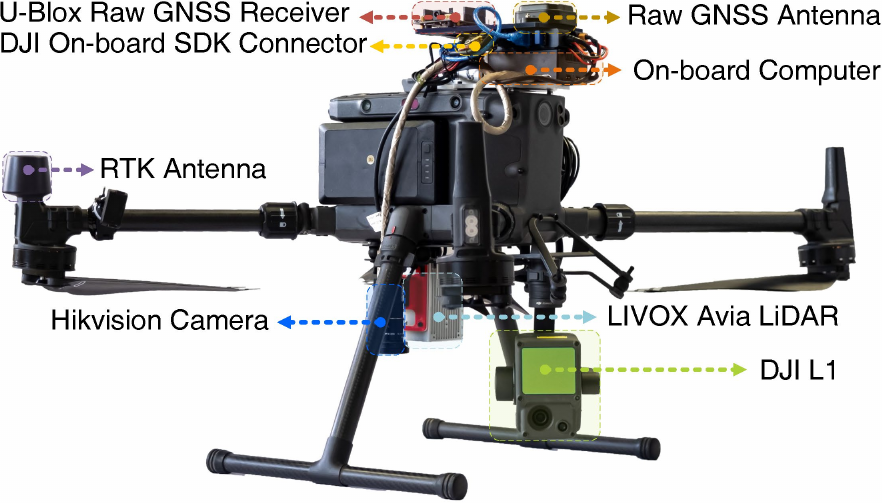}
\includegraphics[width=0.6\linewidth]{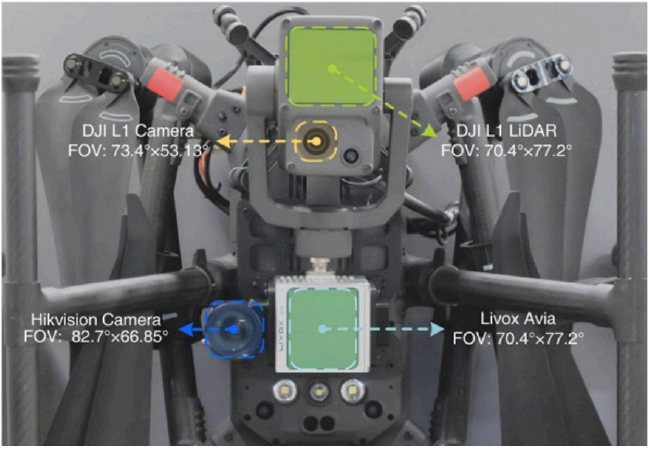}
\vspace{-8pt}
\caption{The data collection platform used in Mars-LVIG. Images are from Mars-LVIG \cite{li2024mars_lvig}. The sensors are mounted on the DJI M300 RTK industrial UAV.}
\vspace{-8pt}
\label{sfig:dji}
\end{figure}

The data collection platform of Mars-LVIG \cite{li2024mars_lvig} is shown in \cref{sfig:dji}, in which sensors are mounted on a DJI M300 RTK industrial UAV. The main perception sensors include a DJI L1 camera, a DJI L1 LiDAR, a Hikvision camera, and a Livox Avia LiDAR. Among them, the DJI L1 LiDAR is a close-form sensor, where frame-wise point clouds are not accessible. The DJI L1 LiDAR can only be processed using DJI Terra. 
In our UAVScenes, we leverage images and frame-wise point clouds produced from the Hikvision camera and the Livox Avia LiDAR, both of which are synchronized at the hardware level with the perception rate at 10Hz.

\subsection{Data Processing}
The original images, LiDAR point clouds, and GNSS coordinates provided by Mars-LVIG are packed and stored in the ROS Bag format. Their topics are:
\begin{itemize}
\item /left\_camera/image/compressed
\item /livox/lidar
\item /dji\_osdk\_ros/rtk\_position
\end{itemize}
We use ROS-Noetic-Ubuntu-20.04 to extract the frame-wise data. The images all have the same size of 2448$\times$2048$\times$3. For the LiDAR point cloud, we have conducted noise filtering to exclude outlier points. For the GNSS coordinates, we apply interpolation to extend them to fit the number of image frames used for Terra SfM reconstruction.
\begin{itemize}
\item DJI Terra. \url{https://enterprise.dji.com/dji-terra}
\end{itemize}

\begin{figure}[!htb]
\centering
\includegraphics[width=0.8\linewidth]{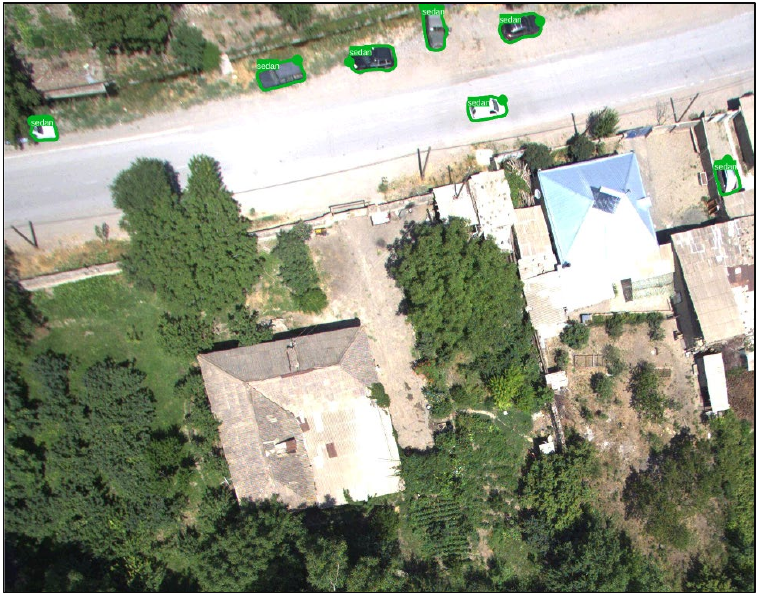}
\vspace{-8pt}
\caption{Image-based annotating for dynamic objects.}
\vspace{-8pt}
\label{sfig:supp_viz_annotating}
\end{figure}

Our labeling work is conducted based on open-source tools X-Anylabeling and CloudCompare.
\begin{itemize}
\item X-Anylabeling. \url{https://github.com/CVHub520/X-AnyLabeling}
\item CloudCompare. \url{https://www.danielgm.net/cc/}
\end{itemize}
The visualization of image instance annotating is demonstrated in \cref{sfig:supp_viz_annotating}.

\begin{figure}
\centering
\includegraphics[width=0.95\linewidth]{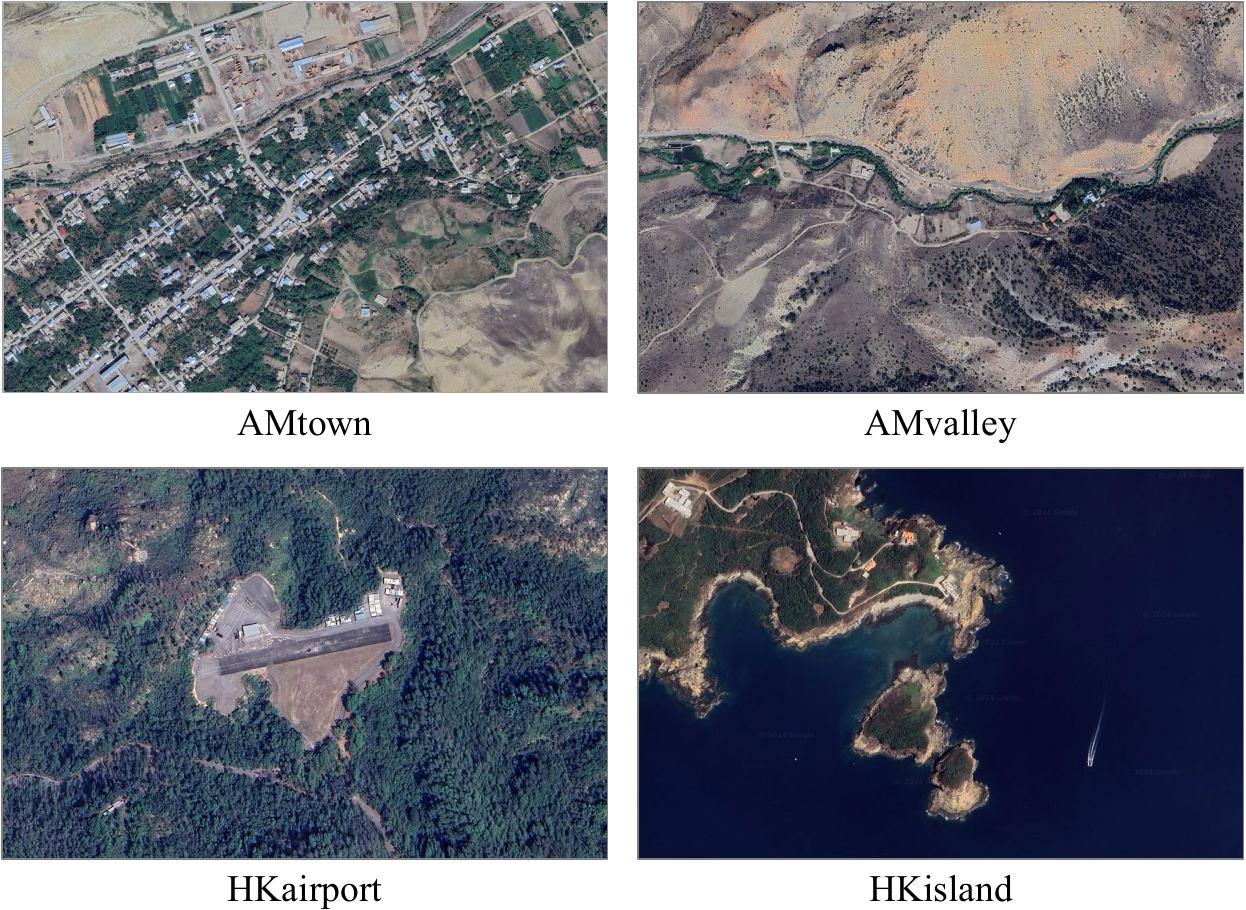}
\vspace{-8pt}
\caption{Visualization of the included scenes under Google Map.}
\vspace{-8pt}
\label{sfig:viz_dataset}
\end{figure}

Our proposed UAVScenes dataset is constructed based on the Mars-LVIG dataset \cite{li2024mars_lvig}. The visualization of some scenes is shown in \cref{sfig:viz_dataset}. We list more details about the dataset as in \cref{stab:sequence_summary}. We consider a total of 23 sequences, where 20 sequences are with full annotations on images and LiDAR point clouds as well as 6-DoF poses. The rest "Featureless\_GNSS" related sequences are with instance annotations for dynamic objects. The 8 splits (20 sequences) used for map and 6-DoF poses reconstruction are also shown in \cref{stab:sequence_summary}.

\begin{table*}[!htb]
\centering
\setlength{\tabcolsep}{0.05cm} 
\scriptsize
\begin{tabular}{c | l | cc c | ccc l}
\toprule
\makecell{Map and Pose \\ Recon. Split ID} & Sequence & \makecell{\#Camera Frames \\ with Anno.} & \makecell{\#LiDAR Frames \\ with Anno.}  & \makecell{\#Camera-LiDAR Paris \\ with 6-Dof Poses} & Duration (s) & \makecell{Cruising \\ Speed (m/s)} & \makecell{Cruising \\ Altitude (m)} & Scenarios \& Characters \\
\midrule
\multirow{3}{*}{Split 1} & AMtown01 & 12.9k & 12.9k & 12.9k & 1354 & 4 & 80 & rural towns \\
&AMtown02 & 6.9k & 6.9k & 6.9k & 749 & 8 & 80 & rural towns \\
&AMtown03 & 5.6k & 5.6k & 5.6k & 619 & 12 & 80 & rural towns \\
\midrule
\multirow{3}{*}{Split 2}&AMvalley01 & 11.3k & 11.3k & 11.3k & 1200 & 4 & 130 & valleys \\
&AMvalley02 & 6.3k & 6.3k & 6.3k & 700 & 8 & 130 & valleys \\
&AMvalley03 & 4.8k & 4.8k & 4.8k & 545 & 12 & 130 & valleys \\
\midrule
\multirow{3}{*}{Split 3}&HKairport01 & 7.2k & 7.2k & 7.2k & 779 & 3 & 80 & aero-model airfield \\
&HKairport02 & 3.9k & 3.9k & 3.9k & 455 & 6 & 80 & aero-model airfield \\
&HKairport03 & 3k & 3k & 3k & 365 & 9 & 80 & aero-model airfield \\
\midrule
\multirow{3}{*}{Split 4}&HKairport\_GNSS01 & 6.8k & 6.8k & 6.8k & 790 & 3 & 80 & aero-model airfield \\
&HKairport\_GNSS02 & 3.7k & 3.7k & 3.7k & 462 & 6 & 80 & aero-model airfield \\
&HKairport\_GNSS03 & 3.1k & 3.1k & 3.1k & 396 & 9 & 80 & aero-model airfield \\
\midrule
\multirow{1}{*}{Split 5}&HKairport\_GNSS\_Evening & 7.1k & 7.1k & 7.1k & 786 & 3 & 80 & aero-model airfield, light changing \\
\midrule
\multirow{3}{*}{Split 6}&HKisland01 & 6.6k & 6.6k & 6.6k & 750 & 3 & 90 & islands \\
&HKisland02 & 3.9k & 3.9k & 3.9k & 469 & 6 & 90 & islands \\
&HKisland03 & 3k & 3k & 3k & 379 & 9 & 90 & islands \\
\midrule
\multirow{3}{*}{Split 7}&HKisland\_GNSS01 & 6.8k & 6.8k & 6.8k & 772 & 3 & 90 & islands \\
&HKisland\_GNSS02 & 3.8k & 3.8k & 3.8k & 465 & 6 & 90 & islands \\
&HKisland\_GNSS03 & 2.7k & 2.7k & 2.7k & 390 & 9 & 90 & islands \\
\midrule
\multirow{1}{*}{Split 8}&HKisland\_GNSS\_Evening & 11.1k & 11.1k & 11.1k & 1337 & 3 & 90 & islands, light changing \\
\midrule
\midrule
\multicolumn{2}{c|}{Total} &  120k  & 120k  &  120k & - & -& -& -\\
\midrule
\midrule
\multirow{3}{*}{No Recon.}&Featureless\_GNSS01 & 12.3k (dynamic-only) & - & - & 1397 & 6 & 80 & aero-model airfield \\
&Featureless\_GNSS02 & 4.7k (dynamic-only) & - & - & 561 & 6 & 80 & aero-model airfield \\
&Featureless\_GNSS03 & 11.8k (dynamic-only) & - & - & 1295 & / & 80 & aero-model airfield, high speed \\
\bottomrule
\end{tabular}
\caption{Summary of sequences and their characteristics. The 8 splits are used for 3D reconstruction, which results in 8 3D maps and corresponding camera poses.}
\label{stab:sequence_summary}
\end{table*}

\subsection{Label Class Definition}
In this subsection, we provide the definition for the classes included in our dataset, which is shown in \cref{stab:definition}.

\begin{table*}[!htb]
\centering
\scriptsize
\begin{tabular}{l|c|l}
\toprule
Class Name & Color  & Definition \\
\midrule
Background & \cellcolor{black} &The object that is not covered in the defined classes.\\
Roof & \cellcolor{roof}  &The roof of the building.\\
Dirt Road & \cellcolor{dirt}  &The motor road that is without any traffic signs or traffic labels. \\
Paved Road & \cellcolor{pavedmotor} &The standard road that is with traffic signs and labels. \\
River & \cellcolor{river} &River or sea.\\
Pool & \cellcolor{pool} &The pool used to store water.\\
Bridge & \cellcolor{bridge} &The normal bridge. \\
Container & \cellcolor{container} &The container is any receptacle or enclosure for holding a product used in storage, packaging, and transportation, including shipping. \\
Airstrip & \cellcolor{airstrip} &The road for airplane landing and taking off. \\
Traffic Barrier & \cellcolor{traffic} &Traffic barriers keep vehicles within their roadway and prevent them from colliding with dangerous obstacles.\\
Green Field & \cellcolor{greenfield} &The green field is the land with grass, trees, or farmlands. \\
Wild Field & \cellcolor{wildfield} &The field that is without or with few plants.\\
Solar Panel & \cellcolor{solar} &The normal solar panel. \\
Umbrella & \cellcolor{umbrella} &The normal umbrella.\\
Transparent Roof & \cellcolor{transparent} &The roof that is with transparent materials.\\
Car Park & \cellcolor{carpark} &The normal car park.\\
Paved Walk & \cellcolor{pavedwalk} &The walk for people.\\
Sedan & \cellcolor{sedan} &Normal home-used vehicles, such as normal sedans, SUVs, MPVs, wagons, and hatchbacks.\\
Truck & \cellcolor{truck} &The truck is a motor vehicle designed primarily to transport cargo.\\
\bottomrule
\end{tabular}
\caption{Label class definition.}
\label{stab:definition}
\end{table*}

\section{Experiment Details}
In this section, we provide more details about the experiment implementations. The code-level instructions (e.g., requirements, running scripts, and train/test splits) are also attached at the anonymous site mentioned in \cref{ssec:dataset_details}.

\subsection{Image Semantic Segmentation}
In image semantic segmentation, we leverage the open-source timm package to conduct evaluations for various backbones, all of which are supported in timm. The input images are all resized and center-cropped to 224$\times$224. The learning rate is set as 0.001, the batch size is set as 16, and we use Adam \cite{kingma2014adam} as the optimizer. 
\begin{itemize}
\item timm. \url{https://github.com/huggingface/pytorch-image-models}
\end{itemize}

\subsection{Frame-Wise LiDAR Semantic Segmentation}
We use 3 open-source baselines in our evaluation. For each baseline network, we use their released training pipeline incorporated in the Pointcept package. The link is as follows:
\begin{itemize}
\item PointCept. \url{https://github.com/Pointcept/Pointcept}
\end{itemize}

\subsection{Place Recognition}
Similar to the implementation of image semantic segmentation models, we use a PR model factory package DVGLB with customization to serve as the code base. We also select other open-source baselines to conduct our comparisons. Hard negative mining methods are applied for all models, which is inherently supported in DVGLB. The learning rates are set as 1e-5 and 1e-4 for images and point clouds respectively. Re-ranking methods are not applied in the comparison.  
\begin{itemize}
\item deep-visual-geo-localization-benchmark (DVGLB). \url{https://github.com/gmberton/deep-visual-geo-localization-benchmark}
\item MixVPR\cite{ali2023mixvpr}. \url{https://github.com/amaralibey/MixVPR}
\item AnyLoc\cite{keetha2023anyloc}. \url{https://github.com/AnyLoc/AnyLoc}
\item SALAD\cite{izquierdo2024salad}. \url{https://github.com/serizba/salad}
\item MinkLoc3D\cite{komorowski2021minkloc3d}. \url{https://github.com/jac99/MinkLoc3D}
\item MinkLoc3D V2\cite{komorowski2022minkloc3dv2}. \url{https://github.com/jac99/MinkLoc3Dv2}
\item MinkLoc++\cite{komorowski2021minkloc++}. \url{https://github.com/jac99/MinkLocMultimodal}
\item AdaFusion\cite{lai2022adafusion}. \url{https://github.com/MetaSLAM/AdaFusion}
\item LCPR\cite{zhou2024lcpr}. \url{https://github.com/ZhouZijie77/LCPR}
\item UMF\cite{umf2024}. \url{https://github.com/DLR-RM/UMF}
\end{itemize}

\subsection{Novel View Synthesis}
We use open-source SOTA novel view synthesis models, whose links are listed as follows:
\begin{itemize}
\item 3DGS\cite{kerbl:hal-04088161}. \url{https://github.com/graphdeco-inria/gaussian-splatting}
\item GaussianPro\cite{GaussianPro}. \url{https://github.com/kcheng1021/GaussianPro}
\item DCGaussian\cite{wang2024dc}. \url{https://github.com/linhanwang/DC-Gaussian}
\item Pixel-GS\cite{zhang2024pixelgs}. \url{https://github.com/zhengzhang01/Pixel-GS}
\item Instant-NGP\cite{mueller2022instant}. \url{https://github.com/nerfstudio-project/nerfstudio}
\end{itemize}
All models are without any fine-tuning on our UAVScenes dataset. The input image size is set as 2448$\times$2048 for all models. No ensemble methods are applied during the evaluation.

\subsection{6-DoF Visual Localization}
We use open-source SOTA scene coordinate regression models, whose links are listed as follows:
\begin{itemize}
\item ACE\cite{brachmann2023ace}. \url{https://github.com/nianticlabs/ace}
\item GLACE\cite{GLACE2024CVPR}. \url{https://github.com/cvg/glace}
\item FocusTune\cite{10483746}. \url{https://github.com/sontung/focus-tune}
\item AtLoc\cite{wang2020atloc}. \url{https://github.com/BingCS/AtLoc}
\item RobustLoc\cite{wang2023robustloc}.  \url{https://github.com/sijieaaa/RobustLoc}
\end{itemize}
The input image size is set as 573$\times$480 for all models. No ensemble methods are applied during the evaluation.

\subsection{Depth Estimation}
We use open-source SOTA zero-shot depth estimation models, whose links are listed as follows:
\begin{itemize}
\item ZoeDepth\cite{bhat2023zoedepth}. \url{https://github.com/isl-org/ZoeDepth}
\item Depth Anything\cite{yang2024depthanything}. \url{https://github.com/LiheYoung/Depth-Anything}
\item Depth Anything V2\cite{yang2024depthanythingv2}. \url{https://github.com/DepthAnything/Depth-Anything-V2}
\item Metric3D\cite{yin2023metric3d}. \url{https://github.com/YvanYin/Metric3D}
\item Marigold\cite{ke2024repurposing_marigold}. \url{https://github.com/prs-eth/Marigold}
\item GeoWizard\cite{fu2025geowizard}. \url{https://github.com/fuxiao0719/GeoWizard}
\end{itemize}
All models are without any fine-tuning on our UAVScenes dataset. The input image size is set as 448$\times$448 for all models. No ensemble methods are applied during the evaluation. The depth evaluation codes are from Monodepth2. 
\begin{itemize}
\item Monodepth2\cite{monodepth2}. \url{https://github.com/nianticlabs/monodepth2}
\end{itemize}
The least-square alignment is not applied to specially evaluate affine-invariant models including Marigold and GeoWizard.

The evaluation metric is following previous works MonoDepth2\cite{monodepth2}. Suppose there are $N$ predicted and ground-truth depth maps (i.e. $d^{\mathrm{pred}}$ and $d^{\mathrm{gt}}$) and with height $H$ and width $W$ to be evaluated, the metrics are computed as follows.

\noindent For $\mathrm{AbsRel}$:
\begin{align}
\mathrm{AbsRel}_j &= \frac{1}{H W} \sum_{i=1}^{H W} \frac{|d_{i,j}^{\mathrm{pred}} - d_{i,j}^{\text{gt}}|}{d_{i,j}^{\mathrm{gt}}}, \\
\mathrm{AbsRel} &= \frac{1}{N} \sum_{j=1}^{N} \mathrm{AbsRel}_j.
\end{align}
For $\mathrm{SqRel}$:
\begin{align}
\mathrm{SqRel}_j &= \frac{1}{H W} \sum_{i=1}^{H W} \frac{(d_{i,j}^{\mathrm{pred}} - d_{i,j}^{\mathrm{gt}})^2}{d_{i,j}^{\mathrm{gt}}}, \\
\mathrm{SqRel} &= \frac{1}{N} \sum_{j=1}^{N} \mathrm{SqRel}_j.
\end{align}
For $\delta_{1.25}$:
\begin{align}
\mathrm{thresh}_{i,j} &= \max \left( \frac{d_{i,j}^{\mathrm{pred}}}{d_{i,j}^{\mathrm{gt}}}, \frac{d_{i,j}^{\mathrm{gt}}}{d_{i,j}^{\mathrm{pred}}} \right), \\
\delta_{1.25,j} &= \frac{1}{H W} \sum_{i=1}^{H W} \mathbf{1} \left( \mathrm{thresh}_{i,j} < 1.25 \right), \\
\delta_{1.25} &= \frac{1}{N} \sum_{j=1}^{N} \delta_{1.25,j}.
\end{align}

\section{More Dataset Information}

\begin{figure*}[!htb]
\centering
\includegraphics[width=0.7\linewidth]{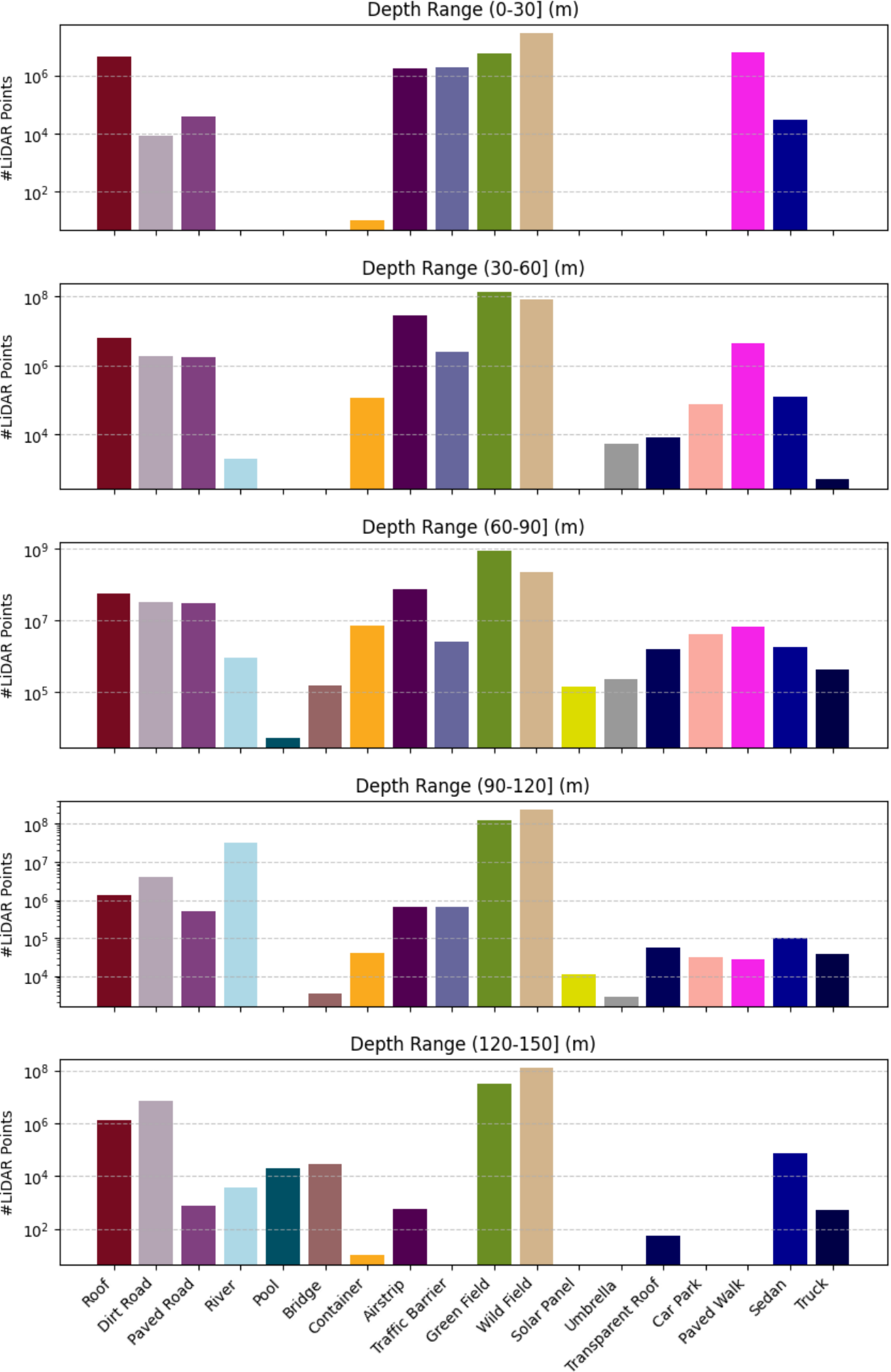}
\vspace{-8pt}
\caption{LiDAR point cloud depth value statistics.}
\vspace{-8pt}
\label{sfig:viz_depth_stats}
\end{figure*}

More visualization of the LiDAR point cloud depth values is in \cref{sfig:viz_depth_stats}.

\begin{figure*}[!htb]
\centering
\includegraphics[width=0.65\linewidth]{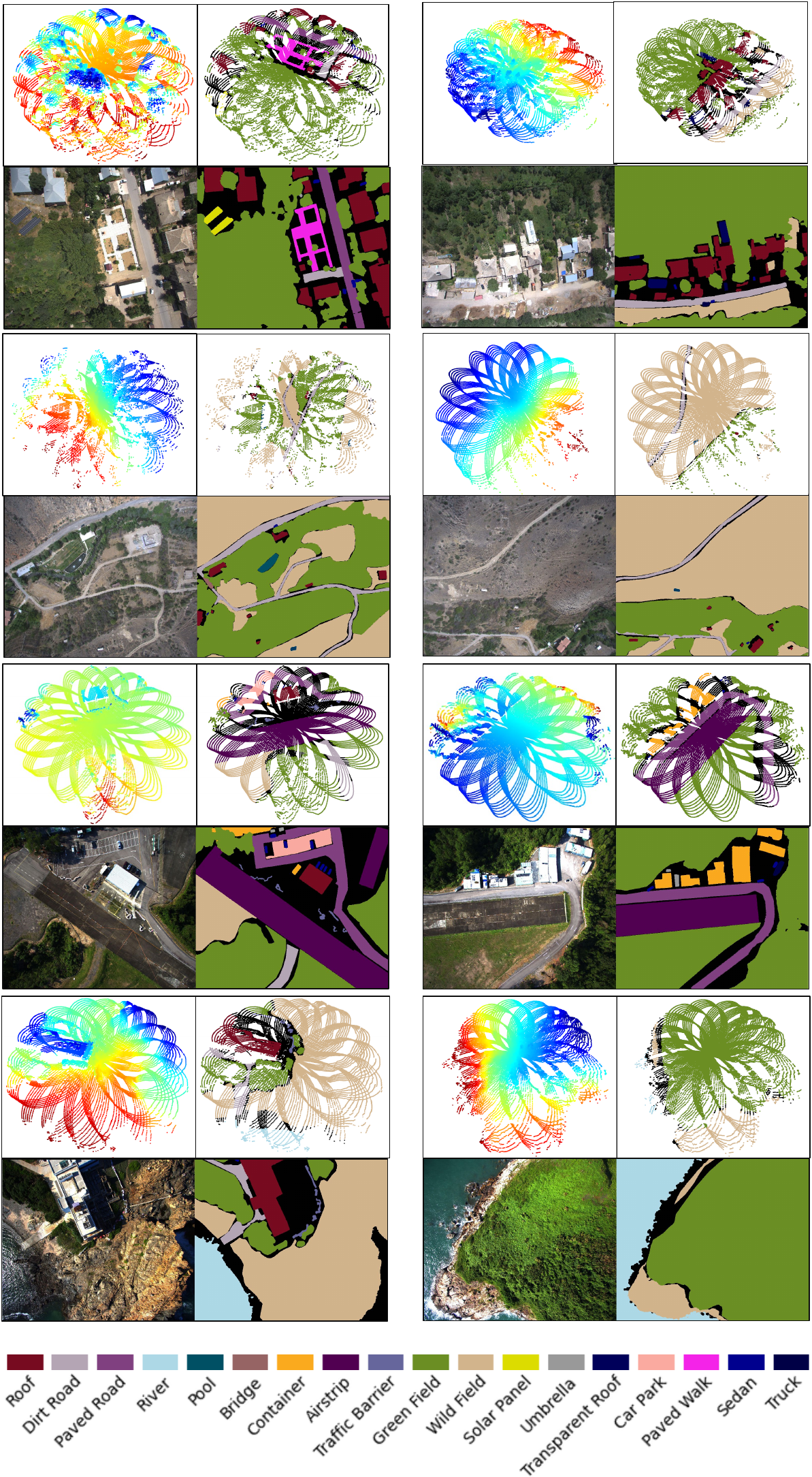}
\vspace{-8pt}
\caption{Visualization of frame-wise image and LiDAR point cloud annotations.}
\vspace{-8pt}
\label{sfig:viz_2d_3d_annotation}
\end{figure*}

More visualization of 2D and 3D semantic annotation examples is in \cref{sfig:viz_2d_3d_annotation}.

\begin{figure*}[!htb]
\centering
\includegraphics[width=0.7\linewidth]{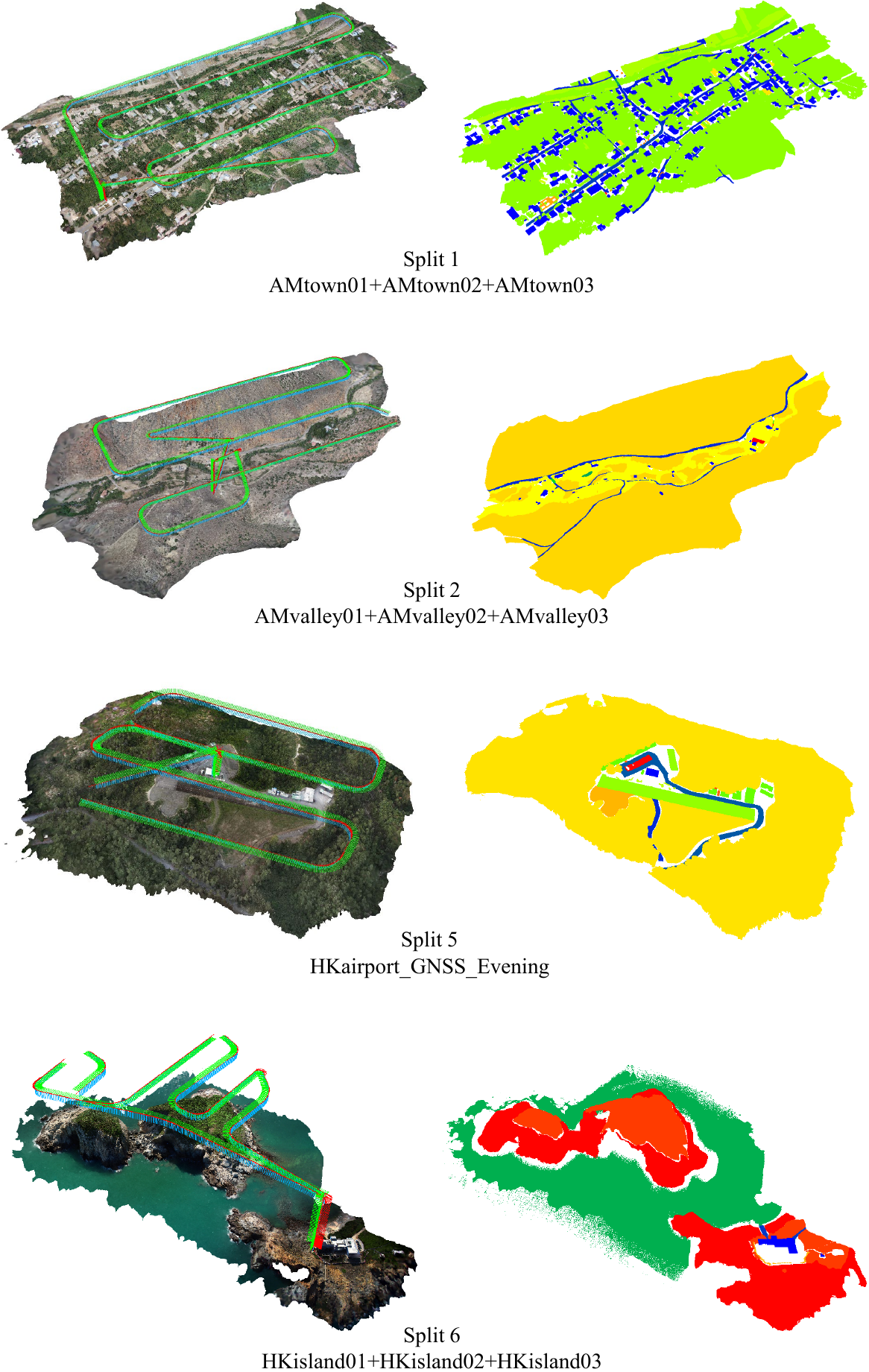}
\caption{Visualization of the annotated 3D map and 6-DoF poses from different sequence splits. The class colors are randomly generated for visualization in CloudCompare.}
\vspace{-8pt}
\label{sfig:viz_3d_maps}
\end{figure*}

More visualization of reconstructed 3D maps and 6-DoF poses is in \cref{sfig:viz_3d_maps}.

{
    \small
    \clearpage
    \clearpage
    \newpage
}

\end{document}